%% file: main.tex
\documentclass{article}
\usepackage[T1]{fontenc}
\usepackage{iclr2027_conference,times}
\usepackage{amsmath,amssymb,amsfonts}
\usepackage{graphicx}
\usepackage{booktabs}
\usepackage{microtype}
\usepackage{xcolor}
\usepackage{url}
\usepackage{algorithm}
\usepackage{algpseudocode}

\newcommand{\method}{\textsc{EviScope}}

\title{\textsc{EviScope}: Evidence-Conditioned Sampling\\for Long-Range 3D Detection}

\author{\input{authors}}

\usepackage[hidelinks]{hyperref}
\iclrfinalcopy
\hypersetup{pdftitle={EviScope: Evidence-Conditioned Sampling for Long-Range 3D Detection},pdfauthor={Mingyang Lyu, Yinqian Sun, Gongce Cao, Chujie Tang, Yuhao Zhang, Lu Jia, Qian Zhang, Bo Wang, Pin Lyu}}
\begin{document}
\maketitle
\lhead{}\rhead{}
\suppressfloats[t]

\begin{abstract}
Query-based camera--radar detectors choose where and over which frames each
object query samples from the query's position or feature. We test whether
radar evidence around the query should also set these choices at the
$150$\,m range of MAN TruckScenes, where recall differs across object-level
radar-return counts within every range band. \method{} is a minimal
test: the current frame's returns form a non-learned bird's-eye
evidence field, which each query reads at its decoded centre to scale its
radial sampling extent and to weight eight frames. It adds $5{,}450$
parameters ($0.009\%$) to RaCFormer, with no latency increase in our timing.
Our module-free recipe exceeds the published TruckScenes mAP ($25.9$ over
seven runs against $24.1$), so we measure the module only against its own
baseline at matched seeds. The evidence a query reads rises with its object's
return count (Spearman $0.51$) but captures zero-return objects only in part,
and the learned radius tracks it ($\rho\approx-0.91$ in four runs). Over seven
four-epoch pairs the module changes NDS by $+0.31\pm0.36$ and mAP by
$-0.27\pm0.22$ (mean $\pm$ SE). Against a geometry-only conditioner with
the same parameter count, trained at three matched seeds, it is higher by
$+1.25\pm0.72$ NDS and $+0.87\pm0.21$ mAP, on mAP in every pair, and the
control's mean NDS is near the baseline's. An exploratory far-range gain failed an internally
pre-specified held-out test, and large inference-time substitution penalties
also occur without query-local evidence: neither learned allocation nor
inference-time dependence substitutes for matched end-to-end comparisons.
\end{abstract}

\section{Introduction}
\label{sec:intro}

\begin{figure}[t]\centering
\includegraphics[width=0.94\linewidth]{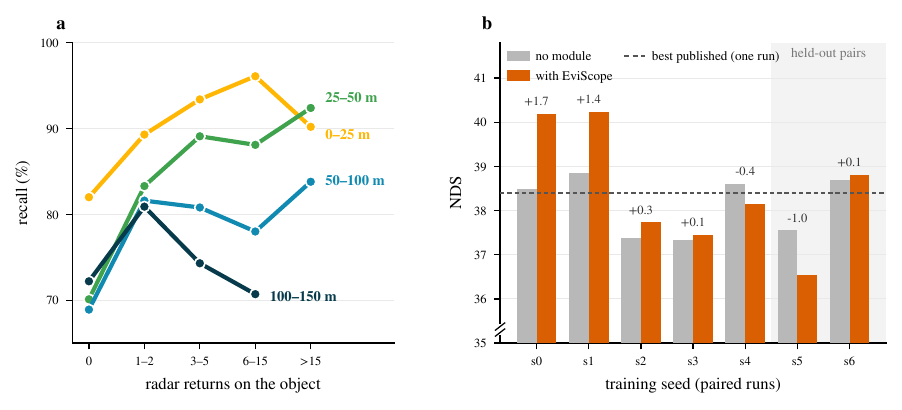}
\caption{\textbf{(a)} Recall of our baseline detector on TruckScenes val by number of
radar returns on the object, one curve per range band (Table~\ref{tab:cross}).
Recall differs across return-count bins within each band, but is not
monotonic in return count; zero-return recall is also not monotonic in range.
\textbf{(b)} NDS of the same detector with and without
\method{}, seven matched seeds, with the paired difference above each pair. The
NDS axis starts at 35.}
\label{fig:teaser}
\end{figure}

Camera--radar fusion combines the semantic detail of vision with radar's range
and velocity measurements for long-range 3D object detection
\citep{truckscenes,centerfusion,crn,rcbevdet}. Detectors of this kind are
query-based: they refine a fixed set of object queries over successive decoder
layers and sample image and bird's-eye-view (BEV) features from each query's
position \citep{detr,detr3d,liu2022petr,sparsebev}. Recent work exploits
geometric structure through concentric query initialisation with adaptive
sampling \citep{racformer}, distance-modulated radar aggregation \citep{sparc},
or object-centred Gaussian representations \citep{horizon3d}. We investigate
whether explicit local radar evidence provides conditioning information for
query sampling beyond query geometry.

Our analysis of the $86{,}252$ evaluated objects of the MAN TruckScenes
validation split \citep{truckscenes} reveals substantial recall differences
across radar-return counts within every range band
(Figure~\ref{fig:teaser}a). Baseline recall increases from $71.4\%$ for
objects with no radar return to $89.6\%$ for those with more than fifteen. Range alone therefore does not fully characterise detection
difficulty, which motivates local radar evidence as an additional sampling cue.

We propose \method{}, an evidence-conditioned query sampling framework that
adapts the spatial sampling extent and the temporal aggregation weights while
keeping the query position as the geometric reference. Current-frame radar
returns form a parameter-free BEV evidence field of multiscale return counts and
radar measurement statistics. Each query reads this field at its decoded centre
and combines the descriptor with its own geometry through a lightweight shared
network. A spatial branch and a temporal branch then predict a radial sampling
scale and a weighting over the eight fused frames, so queries at similar
positions but with different evidence sample differently; a training-only
density branch supervises the shared representation. The change is
small: one non-learned input conditions two sampling controls that RaCFormer
already has. Matched-capacity controls test its contribution;
Table~\ref{tab:inherit} separates inherited and added components. It
adds $5{,}450$ parameters, $0.009\%$ of the detector, with no latency increase
in our timing protocol.

Our experiments separate the performance of the training configuration from the
module's incremental effect. At eight epochs the baseline and \method{} reach
$26.4$ / $39.2$ and $26.7$ / $39.9$ mAP / NDS, both above the strongest published
entry in our comparison ($24.1$ / $38.4$), using a smaller backbone and lower
resolution; at the $256\times864$ input of the published ResNet-50
entry, one \method{} run reaches $28.2$ / $39.9$ against its $23.6$ / $37.4$. These are not
comparisons at matched compute (Section~\ref{sec:recipe}). Across
seven seed-matched pairs at four epochs the module's paired effect is
$+0.31 \pm 0.36$ NDS and $-0.27 \pm 0.22$ mAP (mean $\pm$ SE;
Figure~\ref{fig:teaser}b). Against a geometry-only conditioner with the
same parameter count, at three matched seeds, the module is higher by
$+1.25\pm0.72$ NDS and $+0.87\pm0.21$ mAP, and the control's learned radius
stays at its lower bound for every query (Appendix~\ref{app:review}).
On the seed-0 pair grouped recall rises in $16$ of the $19$ scored
range-by-evidence cells, but the far-range gain tested under a criterion fixed after the first
seeds did not replicate on the held-out pairs
(Sections~\ref{sec:pairs}--\ref{sec:heldout}).

The emitted radial scale tracks local evidence (rank correlations $-0.92$ to
$-0.90$ in four runs), and uniform temporal weights at inference cost
$11.0 \pm 2.0$ NDS (mean $\pm$ SD), but an evidence-blind control also loses
$11.75$ (Section~\ref{sec:analysis}).

\paragraph{Contributions.}
\begin{itemize}\itemsep2pt
  \item A variance-characterised recipe and baseline for camera--radar
        detection on TruckScenes: seven ResNet-50 runs at four epochs average
        $25.9$ mAP (SD $0.45$) against $24.1$ for the best published entry,
        with code in the supplementary material (Section~\ref{sec:recipe}).
  \item \method{} and a controlled test of evidence conditioning: the
        module, geometry-only and shuffled-evidence controls with the same
        parameter count, the density branch at three matched seeds, two
        eight-epoch pairs, and the measured link between object-level and
        query-local evidence (Section~\ref{sec:method}, Appendices~\ref{app:corr}
        and~\ref{app:review}).
  \item Four signals often read as evidence of value, namely an
        evidence-tracking allocation ($\rho\approx-0.91$), $11$-point
        substitution penalties, a $+5$-point subset gain and a $+1.5$ NDS
        two-seed average, all occur here without a replicated four-epoch
        end-to-end gain; matched seeds and held-out tests separate them
        (Sections~\ref{sec:pairs}--\ref{sec:analysis}).
\end{itemize}

\section{Related Work}
\label{sec:related}

\textbf{Long-range camera--radar 3D detection.} Camera--radar fusion builds on
the camera-only BEV lineage \citep{philion2020lss,bevdet,bevdepth,liu2022petr}
and is organised around object-level association, BEV fusion, and sparse
cross-modal interaction. At the object level, \citet{centerfusion} associate
radar detections with image objects through a frustum and \citet{craft}
exchange spatial and contextual cues between proposals with cross-attention. In
BEV, \citet{crn} use radar to assist the view transform and absorb the residual
misalignment with multi-modal deformable attention, while \citet{rcbevdet}
align the branches with a radar-specific encoder and cross-modal attention,
a design \citet{rcbevdetpp} extend to query-based detectors
\citep[see also][]{hydra}. At truck range the cost of dense interaction binds:
4D-radar benchmarks \citep{vod,kradar} annotate shorter ranges, and the
entries that reach $150$\,m on TruckScenes \citep{truckscenes} turn sparse: \citet{sparc} fuse inside sparse
frusta with range-adaptive radar aggregation, and \citet{horizon3d} build a
sparse representation from primitives at estimated keypoints
\citep[see also][]{rctrans}. All of them allocate sampling by position
(Section~\ref{sec:intro}).

\textbf{Modelling the radar return, and reliability-aware fusion.} Radar
returns are uncertain and the two modalities differ in observation quality
\citep{l4dr}. \citet{riccardo} predict the distribution of returns an object
would produce from its class, size and orientation and use it as a post-hoc
matching kernel, which presumes at least one hit. \citet{hgsfusion} densify the points
by modelling direction-of-arrival error. \citet{roburcdet} expand
radar points into Gaussians from cross-section and velocity priors and gate
fusion on camera confidence; \citet{raf} supervise a per-pixel reliability map
from LiDAR--radar--image feature similarity. Both assume a dependable fallback
modality. In our TruckScenes validation analysis, $42.7\%$ of evaluated
objects have no annotated radar return.

\textbf{Query-driven adaptive spatio-temporal sampling.} Query-based detectors
use their observations through spatial sampling and aggregation across frames.
\citet{detr3d} project a 3D reference into the multi-view images;
\citet{bevformer} fuse past BEV features with temporal self-attention;
\citet{sparse4dv2} propagate sparse instance features recurrently instead of
re-sampling every past image. \citet{sparsebev} generate sampling offsets in
space and time from the query feature and mix the sampled features with
predicted weights, and \citet{racformer} combine polar initialisation with
cross-view and cross-modal ray sampling. These policies condition on position
or the query feature; we supply local radar statistics as an explicit,
non-learned input to the spatial scale and the temporal weights, and test what
that input adds beyond geometry (Appendix~\ref{app:review}).

\section{Problem formulation}
\label{sec:problem}

\subsection{Recall by range and radar-return count}
\label{sec:motivation}

For every object the devkit evaluates on TruckScenes val we take the number of
radar returns inside its box, which the dataset annotates, and check whether our
baseline detector (Section~\ref{sec:setup}) places any prediction within
$2$\,m of its centre. We use recall rather than mAP because mAP on a subset of
the ground truth is distorted by false positives that belong to other subsets.

\textbf{Pooled recall differs across return-count bins.}
Table~\ref{tab:cross} in Appendix~\ref{app:crossm} gives the numbers behind
Figure~\ref{fig:teaser}a; its last row pools the ranges. $42.7\%$ of the
evaluated objects receive no return at all, and these are recovered at
$71.4\%$, against $89.6\%$ for objects with more than fifteen returns. The
pooled relation is monotone across the five bins. On the full trainval annotation set,
$774{,}688$ boxes, the zero-return share varies by $57$ points across range for
cars and by $82$ points across classes at $100$--$150$\,m, while the largest
spread across any weather or time-of-day tag in a band is $10$ points
(Appendix~\ref{app:data}). These are descriptive associations, without
adjustment for differences in class or scene composition. At $100$--$150$\,m,
$72.6\%$ of cars and $23.4\%$ of trucks carry no return in the current frame.

\textbf{Differences persist within range bands.} In the nearest band,
zero-return recall is $82.0\%$, versus $89.3$--$96.1\%$ in the nonzero bins.
However, recall is not monotonic in return count within any of the four bands.
At $100$--$150$\,m, for example, recall in the $6$--$15$ bin is $70.7\%$,
below the zero-return value of $72.2\%$. These cross-tabulations motivate
investigating radar evidence, but do not establish that it predicts detection
better than range or adds predictive information after controlling for class,
size and other object attributes.

The motivating statistic uses ground-truth boxes, whereas the module below
reads a local field at a predicted query centre. The quantities can differ
when a query is mislocalised or samples nearby clutter. Appendix~\ref{app:corr} measures their correspondence, which is graded but partial; the object-level association motivates a hypothesis
about query conditioning rather than validating that hypothesis.

\subsection{Evidence-conditioned query sampling}
\label{sec:method}

\begin{figure}[t]\centering
\includegraphics[width=\linewidth]{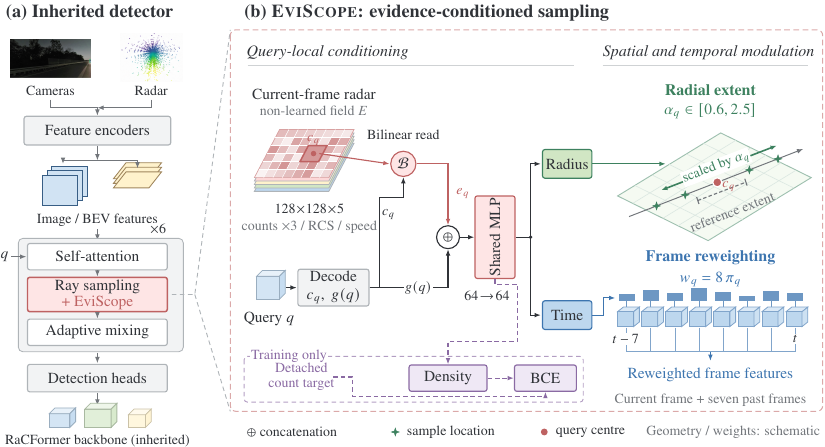}
\caption{(a) The inherited RaCFormer detector, redrawn compactly; \method{}
attaches only to the ray-sampling step of the shared decoder layer.
(b) The module. The current frame's returns form a non-learned field $E$; each
query reads it at its decoded centre $c_q$ ($e_q$) and joins its geometry
$g(q)$; a shared MLP feeds the radial extent $\alpha_q$, the frame weights
$w_q=8\pi_q$ and a training-only density head (dashed). Sample locations and
weights are schematic.}
\label{fig:method}
\end{figure}

We test whether current-frame local radar evidence can guide a query's
sampling extent and temporal aggregation. Sparse evidence motivates a wider
search and a different weighting of past frames, but does not identify which
past frames contain useful returns. We implemented this as a small addition to RaCFormer
\citep{racformer}, shown in Figure~\ref{fig:method}. Everything outside the module, including the polar queries, ray sampling, the multi-frame samplers and the heads, is RaCFormer's; Table~\ref{tab:inherit} separates inherited and added components.

\textbf{Evidence field.} Radar returns for the current keyframe (the five
most recent sweeps of each radar) are scattered into a $128\times128$
bird's-eye grid over the $\pm153.6$\,m detection range. Each cell holds
the log-count of returns in the cell and in its $3{\times}3$ and $5{\times}5$
neighbourhoods, mean radar cross-section and mean speed, giving a five-channel
field $E$. The field has no parameters and is recomputed per sample.

\textbf{Per-query descriptor.} The decoder carries each query as a ray:
polar centre $(\theta,d)$, height, log-size, heading and velocity. We
convert each query to its metric centre $c_q$ and mean predicted box
dimension $s_q$ in the same way the detector's own consumers do, read the field there by bilinear
sampling, $e_q=E(c_q)$, and form, with $\rho_q=\lVert c_{q,xy}\rVert_2$ and $\tilde{\rho}_q=\operatorname{clip}(\rho_q/153.6,0,2)$,
\begin{equation}
d_q=\big[e_q,\,g(q)\big],\quad g(q)=\big[\tilde{\rho}_q,\ (\rho_q{+}1)^{-1},\ s_q/(\rho_q{+}1),\ \log(1{+}s_q)\big],\quad h_q=\mathrm{MLP}(d_q),
\label{eq:desc}
\end{equation}
where the MLP has two ReLU layers of 64 units.

\textbf{Three branches.} Three linear heads read $h_q$:
\begin{equation}
\alpha_q=0.6+1.9\,\sigma(w_\alpha^{\top}h_q+b_\alpha),\quad
\pi_q=\mathrm{softmax}(W_\pi h_q+b_\pi)\in\Delta^{7},\quad
z_q=w_z^{\top}h_q+b_z.
\label{eq:heads}
\end{equation}
The \emph{spatial branch} scales the radial extent of the layer's ray sampling
by $\alpha_q$ \citep[cf.][]{deformabledetr}: the sampling points' radial
offsets, set per layer by RaCFormer, are multiplied by $\alpha_q$. The
\emph{temporal branch} multiplies the detector's aggregation weight for frame
$t$ by $8\pi_{q,t}$, whose mean over the eight frames is one, in the image and
BEV samplers, without renormalising. It
reads current-frame evidence and geometry only, so it learns a policy over
frame offsets rather than measuring the evidence in each historical frame. The
\emph{density branch} is trained with
\begin{equation}
\mathcal{L}=\mathcal{L}_{\mathrm{det}}+0.1\,\mathrm{BCE}\big(z_q,\ \sigma(4(n_q-0.5))\big),\qquad n_q=\operatorname{sg}\big[E_0(c_q)\big],
\label{eq:loss}
\end{equation}
where $E_0$ is the log-count channel and $\operatorname{sg}$ stops the
gradient; the loss covers the queries of the final decoder iteration. The
target is derived from the input counts, so this is an auxiliary regulariser,
not independent supervision; $z_q$ feeds no prediction and acts on the
detector only through the shared trunk. With zero-initialised head weights
and $b_\alpha=\operatorname{logit}(0.4/1.9)$, training starts with
$\alpha_q=1$, uniform frame weights and $z_q=0$, from a model that is
bit-identical to the baseline.

\textbf{A control module.} An early version of \method{} read the field at the
raw polar values $(\theta,d)$ instead of the decoded centre. Because these
values address one fixed sliver of the grid for every query, the field samples
do not represent the radar returns near each decoded centre; the four
geometric terms remain query-dependent. The version trained without complaint and
scored $26.11$ mAP / $39.34$ NDS. We keep it as the \emph{evidence-blind control}
of Section~\ref{sec:substitution}: it has the same branches and the same
per-query temporal weights, but no correctly aligned query-local evidence.
It is an implementation-derived control evaluated at one seed, not a
replicated geometry-only ablation. A separate geometry-only row is reported
under the shortened one-epoch schedule (Appendix~\ref{app:oneep}); the
full-schedule, three-seed geometry-only control is in Appendix~\ref{app:review}.

\textbf{Cost.} One module serves all six shared decoder iterations. It holds $5{,}450$ of the model's $64.0$M parameters, $0.009\%$. Median latency is $965$\,ms against $997$\,ms for the baseline build and peak
memory is $3.811$\,GB for both (Appendix~\ref{app:eff}).

\section{Experiments}
\label{sec:exp}

\subsection{Setup}
\label{sec:setup}

We use MAN TruckScenes \citep{truckscenes} under the official
\texttt{detection\_cvpr\_2024} protocol: 12 classes, centre-distance matching at
$\{0.5,1,2,4\}$\,m, cars scored to $150$\,m, following the protocol inherited
from nuScenes \citep{nuscenes}. We report the validation split. All models use
camera and radar only, with no LiDAR at inference or as depth supervision. The
detector is RaCFormer \citep{racformer} with a ResNet-50 backbone
\citep{resnet} at $256\times704$ input; every row we train uses class-balanced
resampling \citep{cbgs}, eight temporal frames and four epochs on one node of
eight A100 GPUs at an effective batch of eight. We call this configuration
without the module the \emph{baseline}, and the same configuration with the
module \method{}. Where we report several runs, the runs differ only in the
random seed, which sets the initialisation and the data order; the samplers are
not deterministic, so two runs at the same seed would also differ.
Table~\ref{tab:settings} lists the settings of every comparison.

\subsection{The training recipe}
\label{sec:recipe}

\begin{table}[t]\centering\scriptsize
\setlength{\tabcolsep}{2.4pt}
\caption{TruckScenes val. \textbf{L}: LiDAR, \textbf{C}: camera, \textbf{R}: radar;
bold marks the best camera--radar value in a column, the LiDAR row being a
reference. Four-epoch rows are seven-run means (Table~\ref{tab:seeds}),
eight-epoch rows one run each; the margin over the published entries is present
in the module-free baseline rows. \emph{paired $\Delta$} is the within-seed
difference, not the difference of the rounded means: the seven-pair mean
($\pm0.36$ NDS, $\pm0.22$ mAP; Section~\ref{sec:pairs}) and the single
eight-epoch pair. A V2-99 pair is in Appendix~\ref{app:v299}.
$^\ddagger$Error terms not retrieved (Appendix~\ref{app:budget}).
$^\S$One run at the published ResNet-50 input size, without a paired
baseline (Appendix~\ref{app:budget}).}
\label{tab:main}
\begin{tabular}{llllrrrrrrrr}
\toprule
Method & In. & Backbone & Img size & Ep. & NDS$\uparrow$ & mAP$\uparrow$ &
mATE$\downarrow$ & mASE$\downarrow$ & mAOE$\downarrow$ & mAVE$\downarrow$ & mAAE$\downarrow$ \\
\midrule
CenterPoint-V & L   & Voxel & ---              & --- & 35.3 & 22.6 & 0.461 & 0.405 & 0.468 & 3.028 & 0.261 \\
Far3D         & C   & V2-99 & $640\times960$   & --- & 21.4 & 10.7 & 0.883 & 0.507 & 0.671 & 1.352 & 0.338 \\
RCTrans       & C+R & R50   & $256\times864$   & --- & 22.9 & 12.6 & 1.025 & 0.507 & 0.608 & 0.894 & 0.331 \\
CRT-Fusion    & C+R & R50   & $256\times864$   & --- & 28.8 & 16.9 & 0.833 & 0.512 & 0.444 & 0.852 & 0.322 \\
BEVFusion     & C+R & R50   & $256\times864$   & --- & 30.4 & 18.2 & 0.941 & 0.442 & 0.389 & 0.892 & 0.208 \\
SpaRC         & C+R & V2-99 & $640\times960$   & --- & 35.4 & 22.5 & 0.798 & 0.449 & 0.476 & 0.613 & 0.248 \\
Horizon3D     & C+R & R50   & $256\times864$   & 48 & 37.4 & 23.6 & 0.876 & 0.410 & \textbf{0.331} & 0.625 & 0.198 \\
Horizon3D     & C+R & V2-99 & $640\times960$   & 48 & 38.4 & 24.1 & 0.833 & \textbf{0.404} & 0.355 & \textbf{0.561} & 0.208 \\
\addlinespace
Ours (baseline) & C+R & R50 & $256\times704$   & 4 & 38.1 & 25.9 & 0.655 & 0.418 & 0.537 & 0.747 & \textbf{0.123} \\
Ours (\method)  & C+R & R50 & $256\times704$   & 4 & 38.4 & 25.6 & 0.650 & 0.418 & 0.519 & 0.722 & 0.126 \\
\emph{paired $\Delta$} & & & & 4 & $+0.31$ & $-0.27$ & & & & & \\
\addlinespace
Ours (baseline)$^\ddagger$ & C+R & R50 & $256\times704$ & 8 & 39.2 & 26.4 & --- & --- & --- & --- & --- \\
Ours (\method)  & C+R & R50 & $256\times704$   & 8 & \textbf{39.9} & 26.7 & 0.654 & 0.408 & 0.480 & 0.671 & 0.129 \\
\emph{paired $\Delta$} & & & & 8 & $+0.67$ & $+0.26$ & & & & & \\
\addlinespace
Ours (\method)$^\S$ & C+R & R50 & $256\times864$ & 8 & 39.9 & \textbf{28.2} & \textbf{0.599} & 0.412 & 0.568 & 0.719 & 0.127 \\
\bottomrule
\end{tabular}
\end{table}

\begin{table}[t]\centering\scriptsize
\caption{Recipe ablation on the baseline (no module), ResNet-50. The first row is the
seven-run mean of Table~\ref{tab:seeds}; the others are single runs.
$^\dagger$Without class-balanced resampling an epoch has about a quarter the
iterations, so this run was given 12 epochs with the same fraction of training
before the multi-frame phase; it still sees about $30\%$ fewer samples and
differs from the first row in more than one setting.}
\label{tab:recipe}
\begin{tabular}{lrrrrr}
\toprule
configuration & epochs & mAP$\uparrow$ & NDS$\uparrow$ & mAVE$\downarrow$ & runs \\
\midrule
CBGS, 8 frames (baseline)        & 4  & 25.85 & 38.13 & 0.754 & 7 \\
no CBGS, 8 frames$^\dagger$      & 12 & 23.10 & 35.69 & 0.713 & 1 \\
CBGS, 4 frames                   & 4  & 24.02 & 36.24 & 0.844 & 1 \\
CBGS, 2 frames                   & 4  & 22.36 & 33.91 & 0.957 & 1 \\
\bottomrule
\end{tabular}
\end{table}

Our ResNet-50 baseline reaches $25.9$ mAP over seven runs against $24.1$ for
the highest published mAP in our comparison, with a standard deviation of
$0.45$ mAP across our runs. Table~\ref{tab:main} places it among the published
entries on TruckScenes val, as reported by \citet{horizon3d}
\citep{centerpoint,far3d,rctrans,crtfusion,bevfusion,sparc}. In NDS the four-epoch
baseline is at $38.1$ against $38.4$; repeated runs of the published models
would be needed to assess this difference. The error
terms are not uniformly better: translation and attribute errors improve over
the previous best, scale is close, orientation and velocity are worse.

\begin{table}[t]\centering\scriptsize
\setlength{\tabcolsep}{4pt}
\caption{mAP / NDS by disjoint range band, official protocol restricted to each band. Published rows are from
\citet{horizon3d}, Table~9; our rows are the seed-0 models. Bold marks the best
value in each column.}
\label{tab:distance}
\begin{tabular}{lrrrrrrrr}
\toprule
 & \multicolumn{2}{c}{0--25\,m} & \multicolumn{2}{c}{25--50\,m} & \multicolumn{2}{c}{50--100\,m} & \multicolumn{2}{c}{100--150\,m} \\
\cmidrule(lr){2-3}\cmidrule(lr){4-5}\cmidrule(lr){6-7}\cmidrule(lr){8-9}
 & mAP & NDS & mAP & NDS & mAP & NDS & mAP & NDS \\
\midrule
RCTrans         & 19.9 & 28.4 & 12.5 & 24.1 &  7.9 & 19.9 & 6.2 & 15.6 \\
CRT-Fusion      & 26.2 & 34.4 & 16.5 & 27.3 & 11.0 & 26.3 & 7.0 & 11.4 \\
BEVFusion       & 28.8 & 35.7 & 16.8 & 25.2 & 11.7 & 20.1 & 7.4 & 13.9 \\
Horizon3D (R50) & 36.5 & 43.0 & 23.3 & 37.8 & \textbf{17.0} & \textbf{32.7} & 8.9 & 18.4 \\
\addlinespace
Ours (baseline)         & 41.2 & 44.5 & \textbf{25.3} & 39.0 & 15.7 & 29.2 & 8.6 & 18.1 \\
\textbf{Ours (\method)} & \textbf{43.0} & \textbf{47.0} & \textbf{25.3} & \textbf{39.4} & 16.8 & 30.5 & \textbf{9.4} & \textbf{19.2} \\
\bottomrule
\end{tabular}
\end{table}

\textbf{By range.} Table~\ref{tab:distance} restricts the protocol to disjoint
bands. Both seed-0 models exceed the published ResNet-50 entry in the two nearest
bands. At $100$--$150$\,m, the module arm reaches $9.4$ mAP /
$19.2$ NDS against $8.9$ / $18.4$, while the baseline reaches $8.6$ / $18.1$;
the published entry leads at
$50$--$100$\,m. These are single runs.

\textbf{Training budget.} The previous best trains for 48 epochs. Doubling
ours to eight, with every other setting fixed, lifts both arms above it on both
headline metrics: $26.4$ mAP / $39.2$ NDS without the module and $26.7$ /
$39.9$ with it. Epoch counts are not a compute-normalised comparison:
resampling, iterations per epoch, backbones and input sizes differ, and we
do not report matched total iterations or GPU-hours for the published entries.
Against the four-epoch mean the
module arm gains $1.08$ mAP and $1.47$ NDS, with orientation and velocity
errors falling to $0.480$ and $0.671$ (Appendix~\ref{app:budget}). These are
single runs, and the NDS margin ($0.8$--$1.5$) is the size of the spread
between four-epoch runs (Section~\ref{sec:variance}); the remaining gap to the
previous best is in orientation and velocity.

Table~\ref{tab:recipe} examines class-balanced resampling and temporal window
length. The no-CBGS run is $2.75$ mAP below the baseline mean, but also changes
the schedule and sample exposure, so it does not isolate the effect of CBGS.
The four- and two-frame runs are lower by $1.83$ and $3.49$ mAP, respectively,
with velocity error rising from $0.754$ to $0.957$. These single-run results
are consistent with benefits from longer temporal context, as also observed
in camera-only detectors \citep{bevdet4d,solofusion,streampetr}. They do not
establish whether the benefit comes from radar returns, image context or
motion estimation, nor do they quantify each recipe choice's contribution
across seeds.

\subsection{Seven paired runs do not separate the module from zero}
\label{sec:pairs}

Figure~\ref{fig:teaser}b shows the baseline and \method{} trained seven times
each from matched seeds. Two things are visible
at once. The seed moves both models: their NDS across seeds correlates at
$r=0.79$, and the spread across seeds ($1.5$ NDS for the baseline, $3.7$ for
\method{}) is larger than any within-pair difference after the first two seeds.
And the within-pair difference has no consistent sign: the seven paired NDS
differences, printed above each pair in Figure~\ref{fig:teaser}b and tabulated
with the per-seed scores in Appendix~\ref{app:seeds}, average $+0.31$ with a
standard error of $0.36$ ($t=0.87$ on six degrees of freedom), and the paired
mAP difference is $-0.27 \pm 0.22$. The NDS point estimate is smaller than
its standard error.

We report within-seed differences because the seed is a shared source of
variation \citep{picard2021,dodge2020}. With the same seven seeds in both arms,
the mean paired difference equals the difference of means; pairing matters
for the uncertainty estimate. These estimates do not establish an improvement
or equivalence between the models.
Recall, a count over $86{,}252$ objects that does not depend on score
calibration, is a quieter statistic. Its paired difference over the seven seeds is $+0.12 \pm 0.14$ points ($t=0.86$).

The first two pairs alone average about $+1.5$ NDS, substantially above the
seven-pair estimate and illustrating the sensitivity to which runs are reported.

The module starts at identity and can only act once training has moved it
off that point. The reported differences are $-3.98$ NDS under the shortened
one-epoch schedule, $+0.31 \pm 0.36$ at four epochs, and $+0.67$ NDS /
$+0.26$ mAP for one pair at eight. Because the one-epoch schedule changes
multi-frame activation and the budgets have different numbers of runs, these
values do not establish a training-budget trend. We had fixed beforehand that one pair cannot settle this and that only
a difference above $+1.0$ would justify more pairs (Appendix~\ref{app:prereg});
$+0.67$ is recorded as an intermediate result. During review we completed
one further pair at a reviewer's request (Appendix~\ref{app:review}).

\subsection{The far-range recall gain does not survive a held-out test}
\label{sec:heldout}

\begin{table}[htbp]\centering\scriptsize
\caption{Paired recall difference (\method{} minus baseline, points) on objects at
$100$--$150$\,m with $3$--$15$ returns, the subset where the seed-0 pair gained
most (Appendix~\ref{app:gain}). Seeds 0--4 were examined before the test was
written; the internally pre-specified criterion was that both held-out differences be
positive.}
\label{tab:heldout}
\setlength{\tabcolsep}{4pt}
\begin{tabular}{lrrrrrrr}
\toprule
 & s0 & s1 & s2 & s3 & s4 & s5 (held out) & s6 (held out) \\
\midrule
$\Delta$recall & $+4.96$ & $+0.51$ & $-2.64$ & $+3.27$ & $+2.32$ & $-0.67$ & $-0.59$ \\
\bottomrule
\end{tabular}
\end{table}

On the seed-0 pair, the module's recall gain was not spread evenly. It was
$0.6$--$1.4$ points in every hit bin over all ranges, and $4.3$ and $6.2$ points
for objects at $100$--$150$\,m with $3$--$5$ and $6$--$15$ returns
(Appendix~\ref{app:gain}). This was an appealing pattern: distant objects with a
few returns are where a wider search and a longer temporal window should help
most. When we had five pairs, four of them showed a positive difference on this
subset, and the mean was $+1.68$ points.

Because we had chosen the subset after looking at the data
\citep{gelman2014,simmons2011}, we wrote down a confirmation test before
computing recall for the two remaining pairs \citep{nosek2018}. Here
``held out'' refers to training seeds; all pairs use the same validation
objects. The internally recorded protocol is reproduced in
Appendix~\ref{app:prereg}. The
subset was fixed as the two cells merged by count, the matching threshold at
$2$\,m, and the criterion as both held-out differences being positive.
Table~\ref{tab:heldout} gives the result. The held-out pairs give $-0.67$ and $-0.59$. The seven-pair mean is $+1.02 \pm 0.99$. The gain does not satisfy
the pre-specified replication criterion. This limits the interpretation of
the exploratory subset, without establishing a zero population effect.

\section{Analysis}
\label{sec:analysis}

What the module learns, and what an ablation of it measures, are separate
questions. We
ask whether it learned what it was meant to learn, what the standard
attribution test says about it and whether that test can be trusted, and how
much a single run on this benchmark can resolve.

\subsection{The module learns the intended allocation}
\label{sec:mechanism}

\begin{table}[htbp]\centering\scriptsize
\caption{Forward-hook probe on the trained \method{} model (seed 0), last decoder
layer, $270{,}000$ queries over 300 val samples. \emph{radius} multiplies the
layer's sampling extent; \emph{window} is $\exp(\mathrm{H})$ of the temporal
weights, in frames.}
\label{tab:probe}
\begin{tabular}{lrr@{\hskip 18pt}lrr}
\toprule
local evidence & radius & & range & radius & window \\
\midrule
$[0,0.35)$   & 0.734 & & 0--25\,m    & 0.627 & 5.00 \\
$[0.35,0.7)$ & 0.629 & & 25--50\,m   & 0.637 & 4.94 \\
$[0.7,1.1)$  & 0.611 & & 50--100\,m  & 0.663 & 5.22 \\
$\ge 1.1$    & 0.602 & & 100--150\,m & 0.692 & 5.45 \\
             &       & & $>150$\,m   & 0.799 & 5.12 \\
\bottomrule
\end{tabular}
\end{table}

Table~\ref{tab:probe} reports the allocation the trained model makes. The
sampling radius decreases across the local-density bins and increases across
the range bins. The temporal window is similar in the first two range bands,
rises through $100$--$150$\,m, and falls in the final band. The rank correlation
of radius with local evidence is $\rho=-0.91$ and with range $+0.51$; these
marginal correlations do not separate their contributions. The same probe on three further
independently trained runs gives $-0.90$, $-0.92$ and $-0.91$. The ordering is in
place within the first epoch (Appendix~\ref{app:train}). These four probes
confirm an evidence-associated allocation, while the paired comparisons do
not establish an end-to-end benefit. The observed radii fall in $[0.60, 0.84]$, below the
initial value of $1$: the module shrinks the sampling extent overall and adds
an evidence-dependent spread on top of that.

\subsection{Inference-time dependence does not establish incremental value}
\label{sec:substitution}

\begin{table}[htbp]\centering\scriptsize
\caption{Inference-time substitution: each row replaces one modulation on a trained
model and re-evaluates the same weights, with no retraining. The first block is
mean $\pm$ standard deviation over four ResNet-50 \method{} models; the
remaining models are one run each. \emph{recall} is the fraction of evaluated
objects matched within $2$\,m. The further substitutions and the control model
of Section~\ref{sec:substitution} are in Appendix~\ref{app:channel}.}
\label{tab:channel}
\setlength{\tabcolsep}{4pt}
\begin{tabular}{llrrr}
\toprule
model & substitution & $\Delta$mAP & $\Delta$NDS & $\Delta$recall \\
\midrule
\method{}, 4 runs & radial scale $\to$ per-frame mean & $+0.00 \pm 0.11$ & $-0.03 \pm 0.18$ & $+0.06 \pm 0.07$ \\& temporal weights $\to$ uniform    & $-11.4 \pm 3.2$  & $-11.0 \pm 2.0$  & $-8.2 \pm 2.6$ \\
\addlinespace
\method{}, V2-99  & temporal weights $\to$ uniform    & $-13.23$ & $-10.45$ & $-9.81$ \\
evidence-blind control & temporal weights $\to$ uniform & $-8.71$ & $-11.75$ & --- \\
\bottomrule
\end{tabular}
\end{table}

A common way to attribute a model's behaviour to one component is to disable
that component at inference and measure the drop
\citep{zhou2018units,morcos2018,voita2019,michel2019}. We did this for both of the
module's outputs. We take a trained \method{} model and, at inference only,
replace the radial scale by its per-frame mean, by a shuffled copy of itself, or
by a constant, or replace the temporal weights by uniform ones, which is what
the baseline uses. Everything else is untouched, and we verify that each
intervention fires by counting calls and recording the standard deviation of
the radial scale before and after (Appendix~\ref{app:impl}).

\textbf{The two channels look very different under this test.}
Table~\ref{tab:channel} shows the result. Replacing the radial scale by its
per-frame mean changes NDS by $-0.03 \pm 0.18$ and recall by
$+0.06 \pm 0.07$ points across four models (mean $\pm$ SD). Other radial
substitutions also have much smaller observed effects than temporal
substitution (Appendix~\ref{app:channel}).
Making the temporal weights uniform reduces performance in all four models,
by $11.0 \pm 2.0$ NDS (mean $\pm$ SD), $11.4$ mAP and $8.2$ recall points
on average, and by $10.45$ NDS on the V2-99
model, whose end-to-end comparison runs the other way. This contrast measures
dependence on the trained outputs, not the benefit of introducing either branch.

\textbf{Dependence also occurs without query-local evidence.} Two further
models change the reading. The first is the module trained without the density
branch, which removes the auxiliary supervision on the trunk; making its
temporal weights uniform still costs $7.34$ NDS (Appendix~\ref{app:channel}),
so this dependence can arise without density supervision. The second is the evidence-blind control of
Section~\ref{sec:method}, whose descriptor lacks correctly aligned local radar information and
whose spatial branch, we find, has collapsed to a constant (its standard deviation
across queries is $0.000$). Making its temporal weights uniform costs $11.75$
NDS. This control also depends strongly on its temporal weights; its single
run does not establish equivalence to the correctly conditioned module.

\textbf{What the test measures.} The baseline is trained with uniform temporal
weights from the start and reaches $38.13$ NDS over seven runs. A model trained
with learned per-query weights and then evaluated with uniform ones reaches
$25$--$32$. This is consistent with co-adaptation to the learned weights
\citep{hinton2012}; substitution changes the inputs to downstream computation
without allowing it to adapt. It does not isolate the value of the evidence
signal. Related concerns motivate retraining in feature attribution
\citep{hooker2019}, pruning \citep{liu2019pruning} and architecture selection
\citep{dartspt}. Table~\ref{tab:seeds} estimates the complete module's
incremental effect. Isolating the evidence input requires a deliberately
constructed geometry-only conditioner trained under the same full schedule
and matched seeds; Appendix~\ref{app:review} reports it at three seeds.

\textbf{Training-time controls.} The radial channel could also act during
training and be redundant afterwards. Models trained from scratch with every
radial scale fixed at $0.6$, the value the trained module most often emits,
reach $38.02$, $37.63$ and $37.37$ NDS at three seeds, $0.46$, $1.22$ and
$0.02$ below their paired baselines. These controls contain no modulation
module, so they do not isolate radial adaptation while retaining learned
temporal weights; isolating these branches remains an open comparison.
The seed-0 component ablation
(Appendix~\ref{app:ablation}) is also insufficient to establish each branch's
contribution across seeds.

\subsection{A single run cannot resolve one point of NDS}
\label{sec:variance}

Across seeds in the configuration of Section~\ref{sec:setup}, our seven
baseline runs span $1.51$ NDS (standard deviation $0.67$) and our seven
\method{} runs span $3.70$ ($1.39$). The two best published entries differ by
$1.0$ NDS, and every published entry is a single run, which cannot resolve a
difference of that size.

\textbf{Radar degradation.} Re-scoring the seed-0 pair on identical corrupted
inputs tests sensitivity to radar corruption (Appendix~\ref{app:degrade}). Thinning
the returns, uniformly or at long range, leaves the within-pair difference
between $+0.55$ and $+0.70$ mAP at every rate.

The score is concentrated by scene: highway frames reach $38.1$ NDS against
$22$--$25$ for city and terminal, so a val mean mixes regimes that differ by
more than any effect here (Appendix~\ref{app:cond}).

The observed standard deviation is about twice as large for the module arm;
the seed-2 \method{} run is two NDS below seed 0 after the first epoch and
remains lower at the reported checkpoints (Appendix~\ref{app:train}).

With a paired standard deviation of $0.96$ NDS and an effect of $+0.31$, about
$40$ pairs would bring the mean to twice its standard error and about $80$
would give $80\%$ power \citep{card2020}, roughly $1{,}200$ GPU-days
(Appendix~\ref{app:power}); we report the seven pairs as they stand.

\section{Conclusion}
\label{sec:conclusion}

We introduced \method{} to test radar evidence as an additional signal for
query sampling in camera--radar detection. On TruckScenes, the module-free
recipe exceeds the published mAP values in our comparison, while seven
matched pairs leave the module's incremental benefit unresolved:
$+0.31 \pm 0.36$ NDS and $-0.27 \pm 0.22$ mAP (mean $\pm$ SE).
Against a geometry-only conditioner of the same size, \method{} is
higher by $+0.87\pm0.21$ mAP and $+1.25\pm0.72$ NDS over three matched seeds. An exploratory subset gain fails an
internally pre-specified test on two further seeds, and large temporal
substitution penalties also occur without correctly aligned query-local
evidence. These findings support evaluating learned allocation,
inference-time dependence and paired detection gains separately. Evaluation
beyond this benchmark remains necessary to assess the broader hypothesis.

\clearpage
\section*{Ethics Statement}
This work uses the public MAN TruckScenes dataset under its research licence;
it involves no human subjects and no personal data beyond what the dataset
releases with faces and licence plates anonymised. The application domain,
perception for autonomous trucks, is safety-relevant. We report sensitivity
to simulated radar-sector failure (Appendix~\ref{app:degrade}) and note that
the evidence descriptor does not explicitly encode sensor coverage, so that
this limitation can be considered before deployment.

\section*{Reproducibility Statement}
Section~\ref{sec:setup} describes the data splits and evaluation settings.
Appendix~\ref{app:impl} documents data preparation, hyperparameters,
training procedures, and implementation differences between comparison
models. Diagnostic definitions and thresholds appear in
Sections~\ref{sec:motivation}, \ref{sec:mechanism}, and~\ref{sec:substitution},
with component ablations and additional diagnostics in
Appendices~\ref{app:ablation}, \ref{app:oneep}, \ref{app:channel},
and~\ref{app:degrade}. Validation-set results are explicitly identified,
and exploratory and held-out analyses are distinguished in
Section~\ref{sec:heldout} and Appendix~\ref{app:prereg}.
Training and evaluation code is included in the supplementary material
and will be publicly released.

\section*{AI Use Statement}
In this work, we used generative AI tools only to edit author-written text for
grammar, style, and readability. The authors reviewed all AI-assisted revisions
and take full responsibility for the final manuscript, including its claims,
citations, and experimental reporting.

\clearpage
\bibliographystyle{iclr2027_conference}
\bibliography{refs}

\appendix
\section{Internally pre-specified held-out test}
\label{app:prereg}
The far-range low-evidence subset of Section~\ref{sec:heldout} was found on
the seed-0 pair and examined on seeds 1--4. Before recall was computed for the
seed-5 and seed-6 pairs, we fixed the following, reproduced here from our
laboratory log with only the wording translated. The held-out units are
training seeds evaluated on the same validation split; the description here
documents an internal protocol.

\begin{quote}\small
\emph{Origin.} The subset was identified on the seed-0 recall table
(Table~\ref{tab:gain}) and has been examined on seeds 1--4; this is
exploratory. Recall has never been computed for the seed-5 and seed-6 pairs;
they form the held-out set.

\emph{Hypothesis.} \method{} improves recall over the same-seed baseline on
objects at $100$--$150$\,m with $3$--$15$ radar returns, taking the
$3$--$5$ and $6$--$15$ cells merged by object count as one recall rate.

\emph{Criterion.} If the merged recall difference is positive on both held-out
pairs, the result is recorded as replicated; if either is $\le 0$, as not
replicated. The seven-pair estimate (mean $\pm$ SE) is reported as well, but the
conclusion rests on the two held-out pairs only. Overall recall and other cells
are not criteria for this test.

\emph{Not permitted.} Changing the cells, the range boundaries, the matching
threshold (fixed at $2$\,m), or dropping any seed. The result is reported
either way.
\end{quote}

The held-out differences were $-0.67$ and $-0.59$ (Table~\ref{tab:heldout}).
The test was recorded as not replicated.

\paragraph{Training-budget hypothesis.} Before the eight-epoch pair of
Section~\ref{sec:pairs} was trained we recorded the following.

\begin{quote}\small
\emph{Motivation.} The module is initialised at identity and must leave that
initialisation before it can act. The paired difference is monotone in budget:
$-1.66$ mAP / $-3.98$ NDS at one epoch, $+0.31 \pm 0.36$ NDS over seven pairs
at four. The published best trains for 48 epochs; we have four.

\emph{Hypothesis.} Raising the budget from four to eight epochs increases the
paired $\Delta$NDS above $+0.31$.

\emph{Design.} The two eight-epoch runs are the existing four-epoch
configurations of both arms with the epoch count changed from 4 to 8 and
nothing else; same hardware, same batch size per device, same seed (0).

\emph{Criterion.} The eight-epoch paired $\Delta$NDS decides. One pair cannot
settle the question (run-to-run spread $1.5$--$3.7$ NDS). If $\Delta$NDS
$> +1.0$, two further pairs are warranted; if $\le 0$, the budget hypothesis is
recorded as not supported and no further pairs are trained.

\emph{Not permitted.} Changing the epoch count after seeing the result and
reporting selectively; unequal budgets between arms.
\end{quote}

The pair gave $+0.67$ NDS / $+0.26$ mAP, between the two thresholds; it is
recorded as an intermediate result and no claim about the module rests on it.
The motivation above records our hypothesis at the time. The one-epoch
experiment also changes the multi-frame schedule, and only one pair was run
at eight epochs, so these results do not establish a budget-dependent effect.
The additional completed eight-epoch pair of Appendix~\ref{app:review}
was added during review at a reviewer's request, not because this criterion
called for it.

\section{The seven seed-matched pairs}
\label{app:seeds}
\begin{table}[t]\centering\scriptsize
\caption{The ResNet-50 pair at seven seeds, in the configuration of
Section~\ref{sec:setup}; $\Delta$ is \method{} minus baseline within a seed. Seeds 5 and 6 were trained after the held-out test of
Section~\ref{sec:heldout} was written down.}
\label{tab:seeds}
\setlength{\tabcolsep}{2.6pt}
\begin{tabular}{lrrrrrrrr}
\toprule
 & s0 & s1 & s2 & s3 & s4 & s5 & s6 & mean \\
\midrule
\multicolumn{9}{l}{\emph{mAP}} \\
baseline & 25.84 & 26.48 & 26.17 & 25.00 & 25.86 & 25.82 & 25.76 & 25.85 \\
\method  & 26.47 & 26.51 & 25.77 & 24.42 & 25.79 & 24.56 & 25.52 & 25.58 \\
$\Delta$ & $+0.63$ & $+0.03$ & $-0.40$ & $-0.58$ & $-0.07$ & $-1.26$ & $-0.24$ & $-0.27$ \\
\addlinespace
\multicolumn{9}{l}{\emph{NDS}} \\
baseline & 38.48 & 38.85 & 37.39 & 37.34 & 38.60 & 37.55 & 38.69 & 38.13 \\
\method  & 40.19 & 40.24 & 37.73 & 37.44 & 38.16 & 36.54 & 38.80 & 38.44 \\
$\Delta$ & $+1.71$ & $+1.39$ & $+0.34$ & $+0.10$ & $-0.44$ & $-1.01$ & $+0.11$ & $+0.31$ \\
\bottomrule
\end{tabular}
\end{table}

\section{Training budget}
\label{app:budget}
Table~\ref{tab:budget} gives the eight-epoch runs of Table~\ref{tab:main} in
full, next to the seven-run four-epoch means. Both arms are one run. The
baseline was evaluated on four GPUs and the module run on eight, on different
machines; evaluating one trained model under both set-ups differs by about $0.12$ NDS,
with the module's machine reading lower. Error terms for the eight-epoch
baseline were not retrieved at the time of writing. One further run
trains \method{} for eight epochs at $256\times864$, the input size of the
published ResNet-50 entry, and reaches $28.22$ mAP / $39.86$ NDS (mATE
$0.599$, mASE $0.412$, mAOE $0.568$, mAVE $0.719$, mAAE $0.127$). It has no
paired baseline and says nothing about the module. We had predicted that the
larger input would bring orientation error below $0.47$; it did not, and we did
not pursue resolution further. An earlier run at this size changed hardware and
per-GPU batch mid-training and is discarded (Appendix~\ref{app:impl}).

\begin{table}[htbp]\centering\scriptsize
\caption{Four- versus eight-epoch training, ResNet-50 at $256\times704$, with
the other training settings fixed. Four-epoch rows are seven-run means
(Table~\ref{tab:seeds}); eight-epoch rows are single runs evaluated on
different GPU configurations, as described above.}
\label{tab:budget}
\begin{tabular}{llrrrrrrr}
\toprule
model & epochs & mAP$\uparrow$ & NDS$\uparrow$ & mATE$\downarrow$ & mASE$\downarrow$ & mAOE$\downarrow$ & mAVE$\downarrow$ & mAAE$\downarrow$ \\
\midrule
baseline & 4 & 25.85 & 38.13 & 0.655 & 0.418 & 0.537 & 0.747 & 0.123 \\
baseline & 8 & 26.40 & 39.24 & --- & --- & --- & --- & --- \\
\addlinespace
\method  & 4 & 25.58 & 38.44 & 0.650 & 0.418 & 0.519 & 0.722 & 0.126 \\
\method  & 8 & 26.66 & 39.91 & 0.654 & 0.408 & 0.480 & 0.671 & 0.129 \\
\bottomrule
\end{tabular}
\end{table}

\section{The V2-99 pair}
\label{app:v299}
We trained one pair on the V2-99 backbone. The baseline of that pair is our
strongest same-backbone comparison: at $256\times704$ and four epochs it
exceeds the previous best on both headline metrics
(Table~\ref{tab:main}), against a model of the same backbone trained at
$640\times960$ for 48 epochs. Table~\ref{tab:v299} gives both runs. The within-pair
difference, $-0.1$ mAP and $-1.1$ NDS, is one pair and falls inside the
single-run spread of Section~\ref{sec:variance}; the seven-pair estimate of
Section~\ref{sec:pairs} is the measurement we read for the module.

\begin{table}[htbp]\centering\scriptsize
\caption{The V2-99 pair at $256\times704$ and four epochs, seed 0, single runs.
Every other setting is that of Section~\ref{sec:setup}.}
\label{tab:v299}
\begin{tabular}{lrrrrrrr}
\toprule
model & mAP$\uparrow$ & NDS$\uparrow$ & mATE$\downarrow$ & mASE$\downarrow$ & mAOE$\downarrow$ & mAVE$\downarrow$ & mAAE$\downarrow$ \\
\midrule
baseline & 26.7 & 39.0 & 0.643 & 0.410 & 0.506 & 0.752 & 0.124 \\
\method  & 26.6 & 37.9 & 0.651 & 0.410 & 0.530 & 0.814 & 0.135 \\
\bottomrule
\end{tabular}
\end{table}

\section{What it would take to resolve the effect}
\label{app:power}
With a paired standard deviation of $0.96$ NDS and an effect of $+0.31$, about
$40$ pairs would bring the mean to twice its standard error and about $80$ would
give $80\%$ power \citep{card2020}, roughly $1{,}200$ GPU-days. That budget is
out of proportion to the effect, and we report the seven pairs as they stand.

\section{Exploratory recall gain on the seed-0 pair}
\label{app:gain}
Table~\ref{tab:gain} is the table on which the far-range subset was found. It
is the difference between Table~\ref{tab:crossm} and Table~\ref{tab:cross}
for the seed-0 pair, and it is a single pair; Section~\ref{sec:heldout} and
Appendix~\ref{app:prereg} give the outcome of testing its two largest cells on
further pairs.

\begin{table}[htbp]\centering\scriptsize
\caption{\method{} minus baseline recall, in points, by range band and radar-hit
bin, seed-0 pair only, same protocol as Table~\ref{tab:cross}. The $>$15 cell
beyond $100$\,m holds $23$ objects and is omitted. The two bold cells did not
replicate on held-out pairs.}
\label{tab:gain}
\begin{tabular}{lrrrrr}
\toprule
range & 0 hits & 1--2 & 3--5 & 6--15 & $>$15 \\
\midrule
0--25\,m    & $+1.5$ & $-0.6$ & $+0.7$ & $+0.3$ & $+1.7$ \\
25--50\,m   & $-0.6$ & $+1.5$ & $-1.1$ & $+0.7$ & $+0.3$ \\
50--100\,m  & $+1.1$ & $+1.4$ & $+1.2$ & $+2.1$ & $+1.1$ \\
100--150\,m & $+0.8$ & $+0.5$ & $\mathbf{+4.3}$ & $\mathbf{+6.2}$ & --- \\
\addlinespace
all ranges  & $+0.6$ & $+0.9$ & $+0.8$ & $+1.4$ & $+0.9$ \\
\bottomrule
\end{tabular}
\end{table}

\section{Component ablation at four epochs}
\label{app:ablation}
Table~\ref{tab:ablation4} splits the module at the full schedule, at seed 0.
Every row is a single run. The differences among the first three rows are
smaller than the run-to-run spread of Section~\ref{sec:variance}, and we do not
interpret them; we keep the table because dropping single runs after
seeing them would be selective reporting. The
fixed-radius rows are the training-time control discussed in
Section~\ref{sec:substitution}; a third run at seed 2 scored $25.92$ mAP /
$37.37$ NDS.

\begin{table}[htbp]\centering\scriptsize
\setlength{\tabcolsep}{4pt}
\caption{Component ablation at the full four-epoch schedule, seed 0, single
runs. The first three rows are cumulative and share every other setting. The
last two rows are the baseline with every query's sampling extent fixed at
$0.6\times$, the value the trained module most often emits, and no module.
The geometry-only control is \method{} with its five evidence inputs set to zero (Appendix~\ref{app:review}).}
\label{tab:ablation4}
\begin{tabular}{lrrrrrrr}
\toprule
configuration & mAP$\uparrow$ & NDS$\uparrow$ & mATE$\downarrow$ & mASE$\downarrow$ & mAOE$\downarrow$ & mAVE$\downarrow$ & mAAE$\downarrow$ \\
\midrule
baseline, no module         & 25.84 & 38.48 & 0.663 & 0.417 & 0.522 & 0.711 & 0.132 \\
$+$ radius \& window        & 26.46 & 38.93 & 0.662 & 0.418 & 0.490 & 0.750 & 0.110 \\
$+$ density head (\method) & 26.47 & 40.19 & 0.636 & 0.413 & 0.473 & 0.665 & 0.118 \\
\addlinespace
control: geometry-only (evidence inputs zeroed) & 25.35 & 38.57 & 0.675 & 0.428 & 0.490 & 0.703 & 0.115 \\
control: radius fixed at $0.6\times$, no module & 25.69 & 38.02 & 0.660 & 0.426 & 0.503 & 0.780 & 0.114 \\
\phantom{control: }\emph{same, seed 1}                  & 25.45 & 37.63 & 0.676 & 0.434 & 0.549 & 0.715 & 0.136 \\
\bottomrule
\end{tabular}
\end{table}

\section{Single-epoch ablation}
\label{app:oneep}
The single-epoch suite we ran before the four-epoch ablation of
Appendix~\ref{app:ablation}. All rows use one recipe (CBGS, one epoch,
multi-frame from the first step); the column is internally comparable and not
comparable to the four-epoch tables. The full module's difference from the
baseline and the increment from adding the density head reverse sign between
these schedules, while radius plus window improves both reported metrics in
both. These single-run results do not support extrapolating component
ablations from the shortened schedule.

\begin{table}[htbp]\centering\scriptsize
\caption{Component ablation under a shortened single-epoch schedule. All rows
share one recipe (CBGS, one epoch, multi-frame from the first step) and differ
only in the module configuration, so the column is internally comparable; it is
not comparable to Table~\ref{tab:main}, which uses the full four-epoch
schedule. \emph{radius} and \emph{window} are the two modulation paths;
\emph{evidence input} denotes whether the descriptor sees the evidence field or
only its geometric terms.}
\label{tab:ablation}
\begin{tabular}{lccclrr}
\toprule
row & radius & window & evid.\ head & evidence input & mAP & NDS \\
\midrule
baseline (no module) & --- & --- & --- & ---      & 14.20 & 26.65 \\
radius + window      & \checkmark & \checkmark & --- & full & \textbf{14.95} & \textbf{27.58} \\
radius only          & \checkmark & ---        & --- & full & 12.13 & 25.77 \\
window only          & ---        & \checkmark & --- & full & 10.46 & 24.33 \\
full module          & \checkmark & \checkmark & \checkmark & full       & 12.54 & 22.67 \\
full module          & \checkmark & \checkmark & \checkmark & geometry only & 12.84 & 24.74 \\
\bottomrule
\end{tabular}
\end{table}

\begin{table}[htbp]\centering\scriptsize
\setlength{\tabcolsep}{3pt}
\caption{Per-class AP change from adding the module ($\method$ minus baseline,
same recipe), seed-0 pair only. Per-class differences of this size are within
the run-to-run spread of Section~\ref{sec:variance} and we do not read them.
Absolute per-class values are in Appendix~\ref{app:perclass}.}
\label{tab:perclass}
\begin{tabular}{lrrrrrrrrrrrr}
\toprule
 & car & truck & bus & trail. & o.veh. & ped. & moto. & bike & cone & barr. & anim. & sign \\
\midrule
$\Delta$AP & +0.7 & +1.9 & +1.9 & +1.3 & $-4.1$ & +2.1 & $-0.3$ & +1.2 & +2.5 & $-1.3$ & 0.0 & +1.7 \\
\bottomrule
\end{tabular}
\end{table}

\section{Absolute per-class AP}
\label{app:perclass}
\begin{table}[h]\centering\scriptsize
\setlength{\tabcolsep}{3pt}
\caption{Per-class AP (\%) on TruckScenes val for the rows we train. Rows share the recipe of Table~\ref{tab:main}; seed 0 unless stated.}
\begin{tabular}{lrrrrrrrrrrrr}
\toprule
 & car & truck & bus & trail. & o.veh. & ped. & moto. & bike & cone & barr. & anim. & sign \\
\midrule
baseline (R50)              & 57.8 & 37.5 & 0.1 & 49.0 & 11.0 & 10.4 & 46.6 & 12.5 & 43.3 & 2.9 & 0.0 & 39.0 \\
$+$ radius \& window (R50)  & 58.0 & 38.0 & 0.5 & 49.4 & 11.3 & 10.6 & 46.8 & 11.3 & 43.5 & 9.5 & 0.0 & 38.7 \\
\method{} (R50)             & 58.5 & 39.4 & 2.0 & 50.3 &  6.9 & 12.5 & 46.3 & 13.7 & 45.8 & 1.6 & 0.0 & 40.7 \\
baseline (R50, seed 1)      & 58.2 & 39.2 & 5.3 & 51.3 &  9.3 &  9.4 & 48.3 & 12.0 & 41.0 & 4.5 & 0.0 & 39.4 \\
\method{} (R50, seed 1)     & 58.7 & 38.3 & 0.8 & 51.5 & 12.0 &  7.9 & 47.3 & 14.8 & 40.9 & 6.9 & 0.0 & 39.1 \\
baseline (R50, seed 2)      & 58.0 & 37.6 & 1.5 & 51.0 &  8.1 & 10.2 & 43.7 & 10.0 & 40.5 & 15.1 & 0.0 & 38.5 \\
\method{} (R50, seed 2)     & 58.2 & 37.4 & 1.7 & 49.7 &  4.2 & 11.8 & 50.9 & 12.7 & 40.4 & 1.7 & 0.0 & 40.6 \\
baseline (V2-99)            & 58.2 & 38.6 & 0.7 & 51.1 & 10.2 & 12.8 & 49.3 & 13.1 & 44.8 & 1.1 & 0.0 & 40.9 \\
\method{} (V2-99)           & 59.1 & 38.3 & 1.8 & 51.2 &  9.6 & 12.6 & 46.1 & 11.8 & 42.3 & 5.2 & 0.0 & 41.4 \\
\bottomrule
\end{tabular}
\end{table}

\section{Efficiency}
\label{app:eff}
Measured on one A100 at batch one with the full eight-frame pipeline, after
warm-up, over 50 timed iterations; the two builds differ only in whether the
module is attached. Absolute latency includes data preparation on the host and
is not comparable to throughput figures reported elsewhere under other
protocols; the difference between the two rows is the module's cost.
Analytically the module costs $5{,}312$ multiply-adds per query and
decoder iteration, about $29$\,M per sample for $900$ queries and six
iterations, plus one scatter of the radar returns into the $128\times128$
field.
\begin{table}[h]\centering\scriptsize
\begin{tabular}{lrrrr}
\toprule
 & parameters & of which module & median latency (ms) & peak memory (GB) \\
\midrule
baseline & 63{,}998{,}040 & 0     & 997 & 3.811 \\
\method  & 64{,}003{,}490 & 5{,}450 & 965 & 3.811 \\
\bottomrule
\end{tabular}
\end{table}

\section{Implementation details}
\label{app:impl}
\paragraph{Detector.} RaCFormer \citep{racformer} with a ResNet-50 backbone
initialised from nuImages weights, FPN neck, input $256\times704$, $150$ rays
$\times$ $6$ range clusters $=900$ queries, six shared decoder iterations with
per-layer sampling extents $\{0.08, 0.07, 0.06, 0.05, 0.04, 0.03\}$ of the
scene, eight frames (one current, seven past), radar aggregated over five
sweeps per frame. Training uses class-balanced resampling \citep{cbgs}, an
effective batch of $8$, AdamW at $2\times10^{-4}$ with cosine decay, four
epochs, with multi-frame input enabled from the second epoch. The single-epoch
rows of Table~\ref{tab:ablation} enable multi-frame input from the first step
and stop after one epoch.

\paragraph{Module.} Evidence grid $128\times128$ over $\pm153.6$\,m; count
channels at dilations of $0$, $1$ and $2$ cells, plus mean cross-section and
mean speed; descriptor MLP with two hidden layers of $64$; radius scale
$\alpha_q\in[0.6,2.5]$ from a sigmoid; frame weights from a softmax over $8$
frames scaled by $8$; density head trained with binary cross-entropy against
$\sigma(4(n_q-\tau))$ (Equation~\ref{eq:loss}), where $n_q$ is the
detached log-count at the query's centre and $\tau=0.5$ is a fixed constant.
The auxiliary loss is weighted $0.1$ against the detection loss and
covers the final decoder iteration only, because the shared layer overwrites
the earlier iterations' logits. Parameters: $9\cdot64{+}64$ and $64\cdot64{+}64$
in the trunk, $65$, $520$ and $65$ in the three heads, $5{,}450$ in total.

\begin{table}[htbp]\centering\scriptsize
\caption{Inherited and added components. \emph{RaCFormer}: unchanged from
\citet{racformer}. \emph{Adaptation}: changes needed to run on TruckScenes,
shared by baseline and \method{}; not a method claim. \emph{\method{}}: the
module of Section~\ref{sec:method}, the only difference between the paired
rows of Table~\ref{tab:seeds}.}
\label{tab:inherit}
\begin{tabular}{p{0.66\linewidth}l}
\toprule
component & source \\
\midrule
Camera branch: ResNet-50, FPN, LSS depth and BEV view transform & RaCFormer \\
Radar branch: pillar encoder, radar BEV, radar depth & RaCFormer \\
Polar query initialisation ($150$ rays $\times$ $6$ clusters), query denoising & RaCFormer \\
Shared decoder layer ($6$ iterations): ray sampling in image, LSS-BEV and radar BEV; per-frame aggregation weights & RaCFormer \\
Heads, assigner and losses & RaCFormer \\
\addlinespace
Four cameras, $150$\,m range, $256\times256$ BEV grid, TruckScenes class mapping & adaptation \\
Recipe: class-balanced resampling, four (eight) epochs, multi-frame input from epoch two & adaptation \\
\addlinespace
Evidence field $E$ ($5\times128\times128$, no parameters), Equation~\ref{eq:desc} & \method{} \\
Descriptor $d_q$ and $64$-$64$ MLP, shared by the six iterations & \method{} \\
Radial scale $\alpha_q$ on the ray-sampling extent & \method{} \\
Temporal weights $8\pi_q$ on the per-frame aggregation weights & \method{} \\
Density head $z_q$ (unused for detection; removable at inference) and auxiliary loss & \method{} \\
\bottomrule
\end{tabular}
\end{table}

\begin{algorithm}[htbp]\small
\caption{One decoder iteration with \method{}. Lines marked $\triangleright$ are the module; all others are RaCFormer.}
\label{alg:iter}
\begin{algorithmic}[1]
\Require ray queries; image, LSS-BEV and radar-BEV features of $8$ frames; field $E$ built once per sample from current-frame returns
\ForAll{queries $q$}
  \State $(c_q, s_q)\gets$ metric centre and extent of $q$ \Comment{RaCFormer's conversion}
  \State $\triangleright$ $h_q\gets\mathrm{MLP}(d_q)$; $(\alpha_q,\pi_q,z_q)\gets$ heads$(h_q)$ \Comment{Eqs.~\ref{eq:desc}--\ref{eq:heads}}
\EndFor
\State $\triangleright$ scale each query's radial sampling extent by $\alpha_q$
\State sample image, LSS-BEV and radar-BEV features along each ray
\State $\triangleright$ multiply the per-frame aggregation weights of query $q$ by $8\pi_{q,t}$
\State aggregate, refine the queries, predict boxes
\If{training and final iteration} \State $\triangleright$ add $0.1\,\mathrm{BCE}(z_q,\sigma(4(n_q-0.5)))$ \Comment{Eq.~\ref{eq:loss}} \EndIf
\end{algorithmic}
\end{algorithm}

The implementation computes $z_q$ at inference, but detection does not
consume it. Removing this unused head computation leaves the detections unchanged.

\paragraph{Activation recomputation.} The samplers recompute their activations
during the backward pass to save memory. A modulation reached through shared
mutable state is re-read at recomputation time, by which point the shared
decoder layer has overwritten it with the previous iteration's value, so the
backward pass differentiates a function other than the one the forward pass
evaluated. We therefore pass the radial scale and the temporal weights to the
samplers as explicit tensor arguments, which fixes their values for both
passes.

\paragraph{Reading the field at the decoded centre.} The decoder carries
queries in polar form, $[\theta, d]$ normalised to $[0,1]$, and the metric
centre is recovered by two conversions. The evidence-blind control of
Section~\ref{sec:method} sampled the evidence field at the raw polar values
instead, which addresses one fixed sliver of the grid for every query. Its
descriptor therefore lacked correctly aligned query-local radar information,
yet training converged and
the model scored $26.11$ mAP / $39.34$ NDS at seed 0, between the baseline
($38.48$) and the corrected module ($40.19$) at the same seed. When we later
hooked its radial head we found that it emits the same value for every query
(standard deviation $0.000$ across queries), despite retaining geometric
inputs. This observation does not identify why the radial head became constant.
The probe of
Table~\ref{tab:probe} is the check that the corrected module reads what it is
meant to read; we recommend it to anyone reimplementing a module of this kind,
because an evidence-blind one trains without complaint.

\paragraph{Single-node training.} Every row is trained end-to-end on one
node of eight A100 GPUs. We record one run we discarded: a seed-1 model whose
first epoch was trained on Turing GPUs (FP32 matrix multiply) and whose
remaining epochs were resumed on A100 (TF32) scored $37.92$ NDS, against
$40.24$ for the same seed trained on A100 throughout. When we discarded it we read the gap as a numerics change; the seed-2 pair
later showed that a seed alone moves both models by a similar amount
(Table~\ref{tab:seeds}), so the discarded run may have been an unlucky draw
rather than a corrupted one. We keep the rule regardless: no row in this paper
mixes hardware or per-GPU batch size within a run. A second
discarded run, at $256\times864$ input, changed both hardware and per-GPU batch
size at the same resume point and scored $24.29$ / $36.33$; it is comparable
to no row and is not reported.

\paragraph{Probe.} Forward hooks on the last decoder iteration over $300$ val
samples ($270{,}000$ queries), reporting Spearman rank correlation between the
emitted radius scale and, respectively, local return density at the query's
decoded centre and the query's range.

\paragraph{Degradation.} For each val sample the dropout mask is generated
from a random seed derived from the sample token, so the two models in
Table~\ref{tab:degrade} receive identical inputs. \emph{random} drops each
return independently; \emph{far} drops returns beyond $75$\,m at the stated
rate; \emph{sector} disables two of the six radar sensors.

\section{Dataset statistics}
\label{app:data}
Table~\ref{tab:zero} extends the pooled row of Table~\ref{tab:cross} from the val split to the
full trainval annotation set ($774{,}688$ boxes of the twelve official
classes). An object is zero-return when no point of the current radar frame
falls inside its box. The largest spread across weather or time-of-day tags in
any band is $10$ points; across range it is $57$ points for cars, and across
classes at $100$--$150$\,m it is $82$ points. These marginal summaries do not
isolate the effects of weather, class or scene composition.

\begin{table}[h]\centering\scriptsize
\caption{Fraction (\%) of annotated objects with no radar return in the current
frame, trainval split, against range: by class (top), weather tag (middle) and
time-of-day tag (bottom). Cells with fewer than $200$ objects are blank; the
\emph{all classes} row is the count-weighted mean.}
\label{tab:zero}
\begin{tabular}{lrrrr}
\toprule
 & 0--25\,m & 25--50\,m & 50--100\,m & 100--150\,m \\
\midrule
car           & 15.7 & 32.0 & 51.8 & 72.6 \\
truck         &  3.9 &  8.1 & 11.1 & 23.4 \\
bus           &  0.2 &  7.7 & 12.2 & 15.8 \\
trailer       &  2.1 &  6.4 & 10.1 & 23.6 \\
other vehicle &  0.3 &  7.9 &  7.6 & 24.9 \\
pedestrian    & 40.1 & 70.4 & 86.2 & 97.9 \\
motorcycle    &   -- & 37.8 & 78.5 & 93.7 \\
bicycle       & 34.3 & 58.3 & 82.7 &   -- \\
traffic cone  & 74.9 & 89.2 & 94.8 & 97.3 \\
barrier       & 70.9 & 78.5 & 71.9 &   -- \\
animal        &   -- & 70.1 & 80.2 &   -- \\
traffic sign  & 56.3 & 72.3 & 81.4 & 87.2 \\
all classes   & 23.1 & 40.0 & 51.8 & 66.2 \\
\midrule
clear         & 22.5 & 38.9 & 50.3 & 64.2 \\
overcast      & 24.5 & 43.5 & 55.3 & 69.0 \\
rain          & 22.0 & 36.3 & 49.5 & 66.0 \\
snow          & 26.0 & 44.7 & 48.5 & 61.8 \\
fog           & 19.9 & 38.3 & 49.3 & 64.1 \\
\midrule
morning       & 21.7 & 36.9 & 47.7 & 61.8 \\
noon          & 22.7 & 41.4 & 54.0 & 70.0 \\
evening       & 31.7 & 43.0 & 55.7 & 67.7 \\
\bottomrule
\end{tabular}
\end{table}

\section{Performance by scene tag}
\label{app:cond}
TruckScenes attaches scene-level tags for area, weather, lighting and time of
day. Table~\ref{tab:cond} evaluates the \method{} model (seed 0) on the val
frames carrying each tag, under the official protocol restricted to that
subset. We show the tags with at least $100$ val frames; val has no frame
tagged \texttt{night}, and \texttt{fog}, \texttt{snow} and \texttt{tunnel}
carry $40$ frames each (two scenes), too few to report. Scene type dominates:
highway frames score $38.1$ NDS, city and terminal frames $22$--$25$. The
weather tags span $28$--$41$ NDS, but tags are not independent of scene type
(the rain frames come from a different mix of routes), so the table describes
where the model is strong rather than isolating a cause.

\begin{table}[h]\centering\scriptsize
\caption{\method{} (R50, seed 0) on val subsets defined by the official scene
tags; $n$ is the number of val frames carrying the tag.}
\label{tab:cond}
\begin{tabular}{llrrrr}
\toprule
group & tag & $n$ & mAP & NDS & mATE \\
\midrule
all      & --          & 2995 & 26.5 & 40.2 & 0.636 \\
\midrule
area     & highway     & 1960 & 28.0 & 38.1 & 0.645 \\
         & city        &  320 & 12.4 & 24.9 & 0.796 \\
         & terminal    &  320 & 14.3 & 22.7 & 0.849 \\
         & rural       &  315 & 16.9 & 25.2 & 0.790 \\
\midrule
weather  & clear       & 1560 & 24.8 & 35.8 & 0.677 \\
         & overcast    &  915 & 28.8 & 41.0 & 0.620 \\
         & rain        &  400 & 19.5 & 28.0 & 0.795 \\
\midrule
lighting & illuminated & 1955 & 29.4 & 42.1 & 0.629 \\
         & twilight    &  440 & 20.1 & 27.4 & 0.719 \\
         & dark        &  360 & 21.9 & 28.3 & 0.787 \\
         & glare       &  200 & 16.1 & 24.3 & 0.804 \\
\midrule
daytime  & noon        & 1515 & 22.2 & 34.4 & 0.692 \\
         & morning     & 1000 & 34.6 & 42.3 & 0.668 \\
         & evening     &  480 & 22.9 & 32.7 & 0.692 \\
\bottomrule
\end{tabular}
\end{table}

Table~\ref{tab:condcross} crosses the same tags with range. The decline with
range is present under every tag. Rain and glare cost a roughly constant
fraction of the band's mAP at every range (rain 21--35\,\%, glare
34--40\,\% below the all-frames value of the band); city frames lose
42\,\% at $0$--$25$\,m and about 72\,\% beyond.
The night-time and tunnel tags are absent or too small to report.

\begin{table}[h]\centering\scriptsize
\setlength{\tabcolsep}{3.5pt}
\caption{\method{} (R50, seed 0), official protocol restricted to the val
frames carrying each tag and to objects in each disjoint range band; mAP and
NDS (\%). Tags with at least $100$ val frames.}
\label{tab:condcross}
\begin{tabular}{llrrrrrrrr}
\toprule
 & & \multicolumn{2}{c}{0--25\,m} & \multicolumn{2}{c}{25--50\,m} & \multicolumn{2}{c}{50--100\,m} & \multicolumn{2}{c}{100--150\,m} \\
\cmidrule(lr){3-4}\cmidrule(lr){5-6}\cmidrule(lr){7-8}\cmidrule(lr){9-10}
group & tag & mAP & NDS & mAP & NDS & mAP & NDS & mAP & NDS \\
\midrule
all & -- & 43.0 & 47.0 & 25.3 & 39.4 & 16.8 & 30.5 & 9.4 & 19.2 \\
\midrule
area & highway & 42.5 & 43.5 & 35.1 & 43.2 & 20.6 & 31.5 & 10.6 & 19.1 \\
 & city & 24.8 & 29.7 & 6.7 & 14.5 & 5.0 & 13.9 & 2.7 & 8.4 \\
 & terminal & 27.0 & 30.5 & 12.8 & 20.0 & 4.3 & 10.7 & 2.7 & 5.7 \\
 & rural & 39.9 & 38.6 & 16.6 & 24.2 & 7.2 & 17.6 & 4.9 & 12.8 \\
\midrule
weather & clear & 40.0 & 45.4 & 21.5 & 31.7 & 16.2 & 27.2 & 11.3 & 19.8 \\
 & overcast & 48.7 & 50.3 & 38.6 & 45.7 & 20.0 & 33.0 & 9.7 & 18.7 \\
 & rain & 28.0 & 31.4 & 19.4 & 27.6 & 12.5 & 18.6 & 7.4 & 12.7 \\
\midrule
lighting & illuminated & 46.4 & 49.2 & 34.8 & 44.1 & 17.3 & 32.5 & 10.5 & 19.8 \\
 & twilight & 34.7 & 36.1 & 15.0 & 22.9 & 10.9 & 19.3 & 7.5 & 12.7 \\
 & dark & 26.8 & 31.2 & 22.4 & 29.0 & 22.6 & 28.6 & 9.5 & 14.8 \\
 & glare & 26.8 & 30.9 & 16.0 & 23.2 & 10.1 & 16.9 & 6.2 & 11.5 \\
\midrule
daytime & noon & 38.6 & 41.3 & 19.6 & 32.6 & 11.5 & 23.4 & 8.1 & 18.5 \\
 & morning & 44.9 & 44.5 & 31.0 & 36.5 & 28.6 & 34.6 & 15.3 & 20.5 \\
 & evening & 29.3 & 36.7 & 20.1 & 28.3 & 17.4 & 26.6 & 7.7 & 13.7 \\
\bottomrule
\end{tabular}
\end{table}

\section{Training trajectory}
\label{app:train}
\begin{figure}[h]\centering
\includegraphics[width=\linewidth]{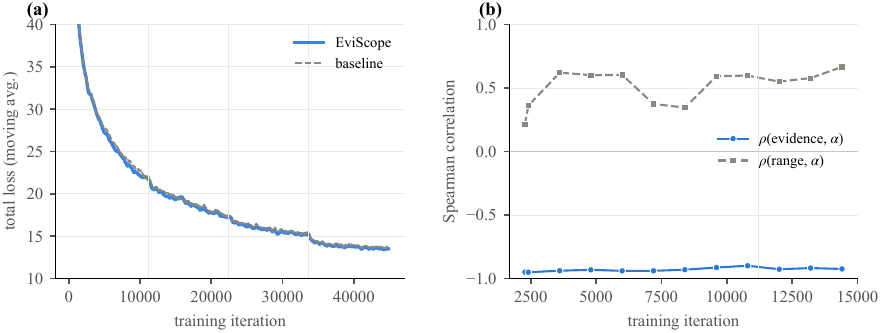}
\caption{(a) Training loss, moving average over 40 iterations, epoch boundaries
marked. (b) Spearman correlation of the emitted radius scale with local
evidence and with range, probed on the \method{} model at intermediate
iterations of the run.}
\label{fig:train}
\end{figure}

Figure~\ref{fig:train}(a) shows the training loss of the \method{} and baseline
runs (seed 0); the module changes neither the level nor the shape of the curve.
Figure~\ref{fig:train}(b) repeats the probe of Table~\ref{tab:probe} on
the model at intermediate iterations over the first $14{,}410$ iterations ($1.3$ epochs). The
correlation between local evidence and radius scale is $\rho=-0.95$ at the first
point probed, iteration $2{,}280$, and stays between $-0.95$ and $-0.90$ over the window;
the final model's value is $-0.91$ (Section~\ref{sec:mechanism}). The range
correlation stays positive and weaker throughout ($0.21$ to $0.67$). The
allocation is not a late-training artefact. Table~\ref{tab:epochs} gives the
val score at the end of epochs 1, 2 and 4 for seed 0 and for seed 2, the low
seed of Table~\ref{tab:seeds}; multi-frame input is enabled from epoch~2. Seed 2
is behind from the first epoch ($29.93$ against $31.88$ NDS) and never closes
the gap at the reported checkpoints. These two trajectories do not establish
when seed-dependent differences become fixed.

\begin{table}[h]\centering\scriptsize
\caption{\method{} (R50) evaluated at the end of training epochs, seeds 0 and 2.}
\label{tab:epochs}
\begin{tabular}{lrrrrrr}
\toprule
 & \multicolumn{3}{c}{seed 0} & \multicolumn{3}{c}{seed 2} \\
\cmidrule(lr){2-4}\cmidrule(lr){5-7}
epoch & 1 & 2 & 4 & 1 & 2 & 4 \\
\midrule
mAP & 20.03 & 23.71 & 26.47 & 17.81 & 21.82 & 25.77 \\
NDS & 31.88 & 36.51 & 40.19 & 29.93 & 34.41 & 37.73 \\
\bottomrule
\end{tabular}
\end{table}

\section{Recall by range and evidence for \method{}}
\label{app:crossm}
\label{app:recall}
\begin{table}[htbp]\centering\scriptsize
\caption{Recall (\%) of our baseline detector (seed 0) on TruckScenes val by range
$\times$ radar-hit count, official $2$\,m matching, no score threshold; object
counts in parentheses, cells below 30 objects omitted. Hit counts are the
official \texttt{num\_radar\_pts} annotation. The last row pools the ranges
over all $86{,}252$ evaluated objects.}
\label{tab:cross}
\begin{tabular}{lrrrrr}
\toprule
range & 0 hits & 1--2 & 3--5 & 6--15 & $>$15 \\
\midrule
0--25\,m    & 82.0 (3{,}885)  & 89.3 (4{,}042) & 93.4 (3{,}455) & 96.1 (4{,}451) & 90.2 (1{,}291) \\
25--50\,m   & 70.1 (9{,}950)  & 83.3 (6{,}512) & 89.1 (3{,}412) & 88.1 (2{,}542) & 92.4 (1{,}341) \\
50--100\,m  & 68.9 (14{,}259) & 81.6 (8{,}688) & 80.8 (3{,}435) & 78.0 (2{,}682) & 83.8 (741) \\
100--150\,m & 72.2 (8{,}758)  & 80.9 (4{,}243) & 74.3 (1{,}643) & 70.7 (899)     & --- \\
\addlinespace
all ranges  & 71.4 (36{,}852) & 83.3 (23{,}485) & 85.9 (11{,}945) & 87.5 (10{,}574) & 89.6 (3{,}396) \\
\bottomrule
\end{tabular}
\end{table}

Table~\ref{tab:cross} reports the baseline. Table~\ref{tab:crossm} gives the
same cross-tabulation for \method{}, with the number of objects in each cell;
Table~\ref{tab:gain} is the difference of the two. Recall rises in $16$ of the
$19$ cells the table reports and the largest decrease is $1.1$ points; the cell
beyond $100$\,m with more than fifteen returns, omitted from the table,
decreases by two objects out of $23$.

\begin{table}[h]\centering\scriptsize
\caption{\method{} recall (\%) by range and by radar returns inside the box
(top), and the number of val objects in each cell (bottom); same protocol as
Table~\ref{tab:cross}.}
\label{tab:crossm}
\begin{tabular}{lrrrrr}
\toprule
 & 0 hits & 1--2 & 3--5 & 6--15 & $>$15 \\
\midrule
0--25\,m & 83.5 & 88.7 & 94.1 & 96.4 & 91.9 \\
25--50\,m & 69.5 & 84.7 & 88.0 & 88.9 & 92.7 \\
50--100\,m & 70.0 & 83.0 & 82.0 & 80.2 & 84.9 \\
100--150\,m & 73.0 & 81.4 & 78.5 & 77.0 & 65.2 \\
\midrule
0--25\,m & 3{,}885 & 4{,}042 & 3{,}455 & 4{,}451 & 1{,}291 \\
25--50\,m & 9{,}950 & 6{,}512 & 3{,}412 & 2{,}542 & 1{,}341 \\
50--100\,m & 14{,}259 & 8{,}688 & 3{,}435 & 2{,}682 & 741 \\
100--150\,m & 8{,}758 & 4{,}243 & 1{,}643 & 899 & 23 \\
\bottomrule
\end{tabular}
\end{table}

\section{Inference-time substitution, per run}
\label{app:channel}
Table~\ref{tab:channelabs} gives the absolute scores behind
Table~\ref{tab:channel}. Each block is one trained \method{} model, the first four ResNet-50 and the
last V2-99; rows within a block share weights and differ only in the
substitution applied at inference.
Evaluations here use four GPUs rather than eight, which moves an absolute score
by up to $0.09$ mAP and $0.05$ NDS against Table~\ref{tab:seeds} and
Section~\ref{sec:method}; every difference within a
block uses the same evaluation configuration. Table~\ref{tab:channelctrl} gives the two control models of
Section~\ref{sec:substitution}; recall was not computed for them, and the mAP
of the radial-mean row was not recorded.

\begin{table}[h]\centering\scriptsize
\setlength{\tabcolsep}{1.8pt}
\caption{Absolute mAP / NDS / recall (\%) for every substitution and run.}
\label{tab:channelabs}
\begin{tabular}{lrrrrrrrrrrrrrrr}
\toprule
 & \multicolumn{3}{c}{run 0} & \multicolumn{3}{c}{run 2} & \multicolumn{3}{c}{run 3} & \multicolumn{3}{c}{run 4} & \multicolumn{3}{c}{V2-99} \\
\cmidrule(lr){2-4}\cmidrule(lr){5-7}\cmidrule(lr){8-10}\cmidrule(lr){11-13}\cmidrule(lr){14-16}
substitution & mAP & NDS & rec. & mAP & NDS & rec. & mAP & NDS & rec. & mAP & NDS & rec. & mAP & NDS & rec. \\
\midrule
none               & 26.48 & 40.18 & 80.10 & 25.77 & 37.76 & 79.51 & 24.38 & 37.39 & 79.64 & 25.79 & 38.11 & 79.29 & 26.55 & 37.92 & 80.19 \\
radial mean        & 26.43 & 40.01 & 80.15 & 25.85 & 37.98 & 79.66 & 24.49 & 37.22 & 79.69 & 25.66 & 38.12 & 79.28 & 26.56 & 37.86 & 80.37 \\
radial shuffled    & 26.44 & 40.06 & 80.02 & 25.70 & 37.75 & 79.59 & 24.43 & 36.92 & 79.50 & 25.70 & 38.04 & 79.28 & 26.53 & 37.74 & 80.21 \\
radial constant    & 26.45 & 40.14 & 80.06 & 25.66 & 37.55 & 79.54 & 24.35 & 37.41 & 79.51 & 25.79 & 38.18 & 79.42 & 26.57 & 37.86 & 80.29 \\
temporal uniform   & 19.60 & 31.93 & 75.66 & 13.68 & 27.11 & 71.06 & 12.07 & 24.91 & 69.40 & 11.37 & 25.70 & 69.54 & 13.32 & 27.47 & 70.38 \\
\bottomrule
\end{tabular}
\end{table}

\begin{table}[h]\centering\scriptsize
\caption{Absolute mAP / NDS for the two control models under inference-time
substitution, seed 0. \emph{no density head}: radial and temporal modulation
trained without the auxiliary density loss. \emph{evidence-blind control}: the early module that
read the evidence field at raw polar coordinates rather than the decoded
query centres (Section~\ref{sec:method}).}
\label{tab:channelctrl}
\begin{tabular}{lrrrr}
\toprule
 & \multicolumn{2}{c}{no density head} & \multicolumn{2}{c}{evidence-blind control} \\
\cmidrule(lr){2-3}\cmidrule(lr){4-5}
substitution & mAP & NDS & mAP & NDS \\
\midrule
none             & 26.41 & 38.91 & 26.02 & 39.30 \\
radial mean      & ---   & 38.81 & ---   & 39.25 \\
temporal uniform & 19.47 & 31.57 & 17.31 & 27.55 \\
\bottomrule
\end{tabular}
\end{table}

\section{Radar degradation}
\label{app:degrade}
\label{sec:robust}

\begin{table}[htbp]\centering\scriptsize
\caption{Paired radar degradation, four-epoch models. Both models are scored on
the same corrupted inputs: the dropout mask is seeded by sample token, so the
comparison is per-point paired rather than per-run. Relative drop is measured
against each model's own clean score, so the two columns are not affected by the
$0.65$ mAP the module already holds when nothing is degraded. \emph{sector
failure} removes a third of the radar sensors.}
\label{tab:degrade}
\begin{tabular}{lrrrrr}
\toprule
 & \multicolumn{2}{c}{mAP} & & \multicolumn{2}{c}{relative drop} \\
\cmidrule(lr){2-3}\cmidrule(lr){5-6}
degradation & \method & baseline & $\Delta$ & \method & baseline \\
\midrule
none                 & 26.48 & 25.83 & $+0.65$ & ---      & ---      \\
random drop $25\%$   & 26.16 & 25.46 & $+0.70$ & $-1.2\%$  & $-1.4\%$  \\
random drop $50\%$   & 24.94 & 24.39 & $+0.55$ & $-5.8\%$  & $-5.6\%$  \\
random drop $75\%$   & 21.11 & 20.42 & $+0.69$ & $-20.3\%$ & $-20.9\%$ \\
far-range drop $50\%$& 26.23 & 25.61 & $+0.61$ & $-0.9\%$  & $-0.8\%$  \\
sector failure $33\%$& 21.72 & 22.00 & $-0.28$ & $-18.0\%$ & $-14.8\%$ \\
\bottomrule
\end{tabular}
\end{table}

Removing radar at evaluation time changes both the detector's features and
the evidence field, so it does not isolate the conditioner.
Table~\ref{tab:degrade} is the seed-0 pair re-scored on
identical corrupted inputs. Thinning the returns, uniformly or preferentially at
long range, leaves the within-pair difference between $+0.55$ and $+0.70$ mAP at
every rate, including one that costs both models a fifth of their accuracy.
Disabling two of the six radar sensors moves it to $-0.28$. In the light of
Table~\ref{tab:seeds} we do not read the sign of a single pair, and the
similarity of the differences under thinning does not establish a robustness
advantage. A possible limitation is that the descriptor contains no explicit
sensor-coverage information: a zero count alone cannot distinguish missing
coverage from an observed region with no returns. Whether this ambiguity
explains the sector-failure result would require a separate control.

\section{Qualitative examples}
\label{app:qual}
\begin{figure}[h]\centering
\includegraphics[width=\linewidth]{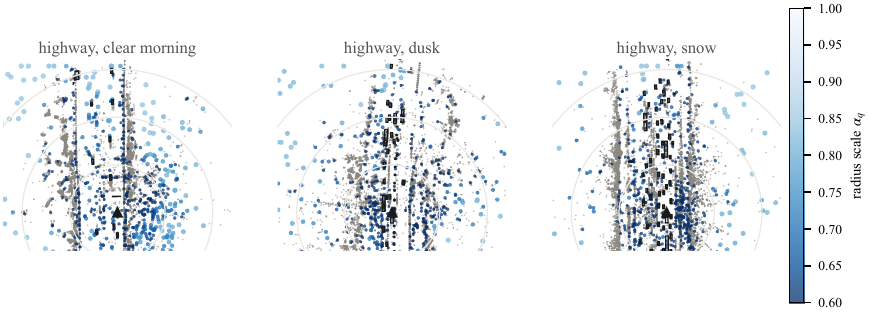}
\caption{Bird's-eye view; ego at the triangle, forward up, rings at $50$, $100$
and $150$\,m. Grey: radar returns. Black: ground-truth boxes. Discs: last-layer
queries at their decoded centres, coloured and sized by the emitted radius
scale $\alpha_q$.}
\label{fig:qual}
\end{figure}

Figure~\ref{fig:qual} shows the radius scale $\alpha_q$ that the trained module
emits for the queries of the last decoder layer on three val frames. Queries
that land on the carriageway, where returns are dense, contract (dark, small);
queries in sparse regions beside the road receive relatively larger scales
(light, large). The observed scales remain below the baseline value of one.
The pattern is the
same in clear, dusk and snow frames.

\section{Query-local evidence and object-level return counts}
\label{app:corr}
Section~\ref{sec:motivation} counts radar returns inside ground-truth
boxes, while the module reads the field at decoded query centres. To measure
how far the second tracks the first, we hooked the last decoder iteration on
$300$ val samples spaced evenly over the split, read the five field channels
at each query's centre with the module's own sampling function, and matched
every ground-truth box within $150$\,m to the nearest unmatched query whose
centre lies within $2$\,m (greedy, one-to-one). Boxes are taken before the
evaluator's point filter, so $47.2\%$ have no annotated return, against
$42.7\%$ of evaluated objects in Section~\ref{sec:motivation}.
Table~\ref{tab:corr} reports two trained models.

\begin{figure}[htbp]\centering
\includegraphics[width=\linewidth]{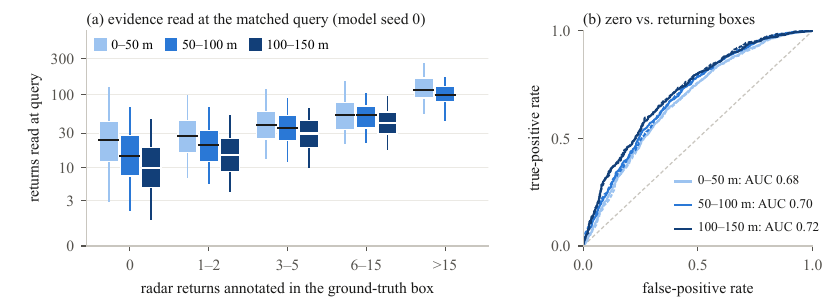}
\caption{(a) Returns read at the matched query's centre (one-cell
dilation, log axis) against the returns annotated in the ground-truth box, by
range band; boxes span the interquartile range, the bar is the median, whiskers
the $5$th--$95$th percentiles; model seed 0, $8{,}277$ matched boxes; cells with
fewer than $20$ boxes are omitted. (b) ROC of the same reading for separating
zero-return from returning boxes, per band; solid seed 0, dashed seed 2.}
\label{fig:corr}
\end{figure}

The query-local reading rises monotonically with the object's return
count. Spearman correlation is $0.51$ at the one-cell dilation and
$0.42$--$0.47$ within each range band, and the reading separates zero-return
from returning objects with AUC $0.72$ overall and $0.67$--$0.72$ within
bands; the two checkpoints agree to within $0.03$. The correspondence is partial. Zero-return objects sit on a
median one-cell reading of $15$ returns, and only $20\%$ of them read below
half a return in the centre cell: the $2.4$\,m cell and its neighbourhood collect
clutter and returns from adjacent objects. The descriptor therefore carries a
graded, noisy proxy of an object's evidence, not the zero-return condition used
to motivate the method. Figure~\ref{fig:corr} shows the full distributions.

\textbf{Where the gap comes from.} The field has no parameters, so it can
also be read at the ground-truth box centres, without a model and for all
$9{,}684$ boxes (bottom of Table~\ref{tab:corr}). Built from the keyframe sweep
alone, the sweep the annotation counts, the field reads below half a return in
the centre cell for $58.3\%$ of zero-return boxes; built as the module builds
it, from the five most recent sweeps of each radar, for $24.4\%$; read at the
matched query centre, for $20.2\%$. Rank correlation with the box's count is
$0.55$, $0.52$ and $0.51$, and AUC $0.76$, $0.73$ and $0.72$. The partial
correspondence therefore comes mostly from how the field is built (five
aggregated sweeps and $2.4$\,m cells read bilinearly) and only slightly from
where a matched query sits relative to its box centre. As a check of the single-sweep reconstruction, every box
with an annotated return has a keyframe return within half its diagonal.
Whether a single-sweep field would condition sampling better is untested.

\begin{table}[htbp]\centering\scriptsize
\setlength{\tabcolsep}{3.5pt}
\caption{Ground-truth return counts against the evidence read at the
matched query's centre, last decoder iteration, $300$ val samples, two
\method{} models (seed 0 / seed 2). $\rho$: Spearman correlation between the
box's return count and the log-count at dilation $0$, $1$, $2$ cells.
AUC: separation of zero-return from returning boxes by the one-cell reading.
Coverage: share of boxes with a query centre within $2$\,m. Middle: median
one-cell reading and share of zero centre-cell readings by the box's return
count, seed 0. Bottom: the non-learned field read at the true box
centres, built from the keyframe sweep that the annotation counts or from the
five sweeps per radar the module uses; \emph{zero read} is the share of
zero-return boxes reading below half a return in the centre cell.}
\label{tab:corr}
\begin{tabular}{lrrrrrr}
\toprule
range & boxes & coverage & $\rho_0$ & $\rho_1$ & $\rho_2$ & AUC$_1$ \\
\midrule
$0$--$150$\,m   & $9{,}684$ & $0.855$ / $0.840$ & $0.48$ / $0.48$ & $0.51$ / $0.51$ & $0.46$ / $0.46$ & $0.72$ / $0.72$ \\
$0$--$50$\,m    & $4{,}058$ & $0.925$ / $0.914$ & $0.45$ / $0.45$ & $0.47$ / $0.47$ & $0.38$ / $0.38$ & $0.68$ / $0.67$ \\
$50$--$100$\,m  & $3{,}458$ & $0.820$ / $0.802$ & $0.42$ / $0.42$ & $0.46$ / $0.46$ & $0.39$ / $0.40$ & $0.70$ / $0.70$ \\
$100$--$150$\,m & $2{,}168$ & $0.779$ / $0.763$ & $0.33$ / $0.35$ & $0.42$ / $0.43$ & $0.40$ / $0.41$ & $0.72$ / $0.72$ \\
\midrule
returns in box & $0$ & $1$--$2$ & $3$--$5$ & $6$--$15$ & $>15$ & \\
median one-cell reading & $15.4$ & $21.9$ & $36.7$ & $51.4$ & $111.2$ & \\
zero centre-cell reading & $20.2\%$ & $8.1\%$ & $2.4\%$ & $0.9\%$ & $0.0\%$ & \\
\midrule
field built from, read at & boxes & & & $\rho_1$ & zero read & AUC$_1$ \\
keyframe sweep, box centre & $9{,}684$ & & & $0.55$ & $58.3\%$ & $0.76$ \\
five sweeps (as used), box centre & $9{,}684$ & & & $0.52$ & $24.4\%$ & $0.73$ \\
five sweeps (as used), matched query & $8{,}277$ & & & $0.51$ & $20.2\%$ & $0.72$ \\
\bottomrule
\end{tabular}
\end{table}

\section{Full-schedule control experiments}
\label{app:review}
This appendix reports completed geometry-only and shuffled-evidence
controls, a multi-seed density-head comparison, and two pairs at eight epochs.
Each model is trained on one node of eight A100 GPUs with a per-GPU batch of
one; cross-node evaluations are marked $\dagger$ in Table~\ref{tab:review}.
Seeds $0$--$2$ are the first three seeds of Table~\ref{tab:seeds}; they include
the two highest-NDS \method{} runs in that seven-seed set.

\textbf{Geometry-only conditioner.} The module of
Section~\ref{sec:method} with the five field channels of $d_q$ replaced by
zeros. It keeps the four geometric terms, the capacity, both learned branches,
the identity initialisation and the auxiliary target computed from the true
counts, so the only difference from \method{} is whether the descriptor sees
local evidence. Four epochs, seeds $0$--$2$, paired with \method{} and the
baseline at the same seeds.

\textbf{Density head across seeds.} \method{} against the same module
trained without the density branch, four epochs.

\textbf{Eight-epoch pairs.} \method{} against the baseline at eight
epochs, with the completed seed-1 pair added to the seed-0 pair of
Table~\ref{tab:budget}. Both seed-1 arms are trained and evaluated on the same
node, unlike the cross-node evaluation of the seed-0 pair.

\begin{figure}[htbp]\centering
\includegraphics[width=\linewidth]{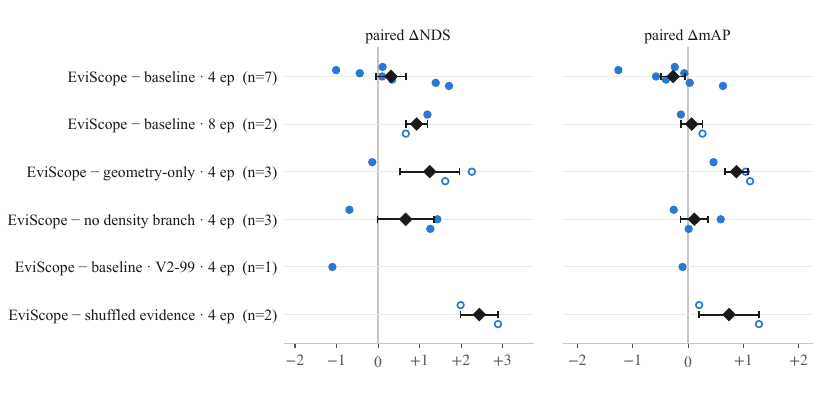}
\caption{Completed within-seed comparisons, first model minus second.
Dots are seeds; hollow dots mark arms evaluated on different nodes
($^\dagger$ in Table~\ref{tab:review}). Diamonds and bars are mean $\pm$ SE;
$n$ counts completed pairs. The four-epoch baseline and V2-99 comparisons,
eight-epoch seed-0 pair and density-head seed-0 comparison are also reported in
Tables~\ref{tab:seeds}, \ref{tab:budget} and \ref{tab:ablation4}, respectively.}
\label{fig:pairs}
\end{figure}

Figure~\ref{fig:pairs} places the completed main within-seed comparisons
on one axis.

\begin{table}[htbp]\centering\scriptsize
\caption{Full-schedule controls; mAP / NDS (\%).
Superscripts identify results reproduced from other tables for paired reference
($^{\mathrm{a}}$Table~\ref{tab:seeds};
$^{\mathrm{b}}$seed-0 row $+$radius \& window of Table~\ref{tab:ablation4};
$^{\mathrm{c}}$seed-0 pair of Table~\ref{tab:budget}).
Only completed comparisons are reported; a dash denotes no reported pair.
$\Delta$ is \method{} minus the row's model within a seed; means use the available pairs.
$^\dagger$Arms evaluated on different nodes (Appendix~\ref{app:budget}); on the
control's node, \method{} scores $26.50$ / $40.19$ (seed 0) and $26.59$ / $40.29$
(seed 1). $^\ddagger$\method{} with evidence inputs permuted at random across the queries of each sample in training and evaluation.}
\label{tab:review}
\begin{tabular}{llrrrr}
\toprule
comparison & model & seed 0 & seed 1 & seed 2 & mean $\Delta$NDS \\
\midrule
\multicolumn{6}{l}{\emph{four epochs}} \\
& \method{}$^{\mathrm{a}}$  & $26.47$ / $40.19$ & $26.51$ / $40.24$ & $25.77$ / $37.73$ & \\
& baseline$^{\mathrm{a}}$   & $25.84$ / $38.48$ & $26.48$ / $38.85$ & $26.17$ / $37.39$ & \\
geometry-only & control & $25.35$ / $38.57^\dagger$ & $25.47$ / $37.98^\dagger$ & $25.31$ / $37.87$ & $+1.25$ \\
shuffled evidence & control$^\ddagger$ & $25.19$ / $37.30^\dagger$ & $26.31$ / $38.25^\dagger$ & -- & $+2.44$ \\
density head  & no density branch & $26.46$ / $38.93^{\mathrm{b}}$ & $25.92$ / $38.81$ & $26.03$ / $38.42$ & $+0.67$ \\
\addlinespace
\multicolumn{6}{l}{\emph{eight epochs}} \\
& \method{}            & $26.66$ / $39.91^{\mathrm{c}\dagger}$ & $26.49$ / $40.08$ & -- & \\
& baseline             & $26.40$ / $39.24^{\mathrm{c}\dagger}$ & $26.62$ / $38.89$ & -- & $+0.93$ \\
\bottomrule
\end{tabular}
\end{table}

\textbf{Geometry-only result.} Over seeds $0$--$2$, \method{} exceeds the
geometry-only control by $+0.87\pm0.21$ mAP and $+1.25\pm0.72$ NDS (mean $\pm$
SE; Table~\ref{tab:review}). mAP is higher in all three pairs ($+1.12$, $+1.04$,
$+0.46$) and translation error is lower in all three (mATE $0.675\to0.636$,
$0.657\to0.624$, $0.662\to0.648$). The NDS differences are $+1.62$, $+2.26$
and $-0.14$. Orientation and velocity error are lower at seeds $0$ and $1$ and
higher at seed $2$ (mAOE $0.490\to0.473$, $0.510\to0.499$, $0.526\to0.545$;
mAVE $0.703\to0.665$, $0.751\to0.641$, $0.744\to0.768$); these two terms
account for the sign of the seed-2 NDS difference. The observed control mean NDS is
$38.14$ against the baseline's $38.24$ and \method{}'s $39.39$, and its mAP is
below the baseline's in all three pairs ($-0.49$, $-1.01$, $-0.86$).
Re-evaluated on the control's node across checkpoints, the three-seed difference
progresses from $+0.11\pm0.46$ mAP / $+0.20\pm0.76$ NDS at epoch $2$ to
$+0.33\pm0.15$ / $+0.55\pm0.50$ at epoch $3$ and $+0.91\pm0.22$ / $+1.26\pm0.73$
at epoch $4$. The per-seed NDS ordering is not uniform: differences for seeds
$0$--$2$ are $(+1.66,-0.18,-0.87)$ at epoch $2$, $(<+0.01,+1.55,+0.10)$ at
epoch $3$, and $(+1.62,+2.31,-0.14)$ at epoch $4$. Across the $75$ validation
scenes, $1{,}000\times$ scene bootstrap
on the three-seed mean gives $95\%$ confidence intervals of $[+0.23,+1.72]$ mAP
and $[-0.28,+2.48]$ NDS ($[-0.08,+2.38]$ mAP and $[-1.05,+3.51]$ NDS under
two-level scene-and-seed bootstrap over $n{=}3$ seeds). In $1{,}000$ random
scene splits, the seed-mean $\Delta$NDS is positive in $94.2\%$ of halves and
in both halves of $88.3\%$ of splits. In the $100$--$150$\,m band, seeds $0$--$2$
give $\Delta$mAP of $+0.7$, $+0.1$, $-0.5$ and $\Delta$NDS of $+1.1$, $+1.2$,
$-0.3$. Thus the NDS intervals
include zero, and neither the checkpoint nor the far-range ordering holds
uniformly across seeds.

\textbf{By scene condition.} Table~\ref{tab:conditions} splits each pair by
the official TruckScenes scene tags, keeping the $20$ conditions with at least
$300$ validation frames, and scores each subset with the official metrics on
its frames. The seed-mean difference is positive in $19$ of the $20$
conditions; the exception is rural scenes ($-0.11$ NDS).

\begin{table}[htbp]\centering\scriptsize
\setlength{\tabcolsep}{4pt}
\caption{\method{} minus the geometry-only control by official scene
condition, $\Delta$NDS (points) at every seed. Conditions with at least $300$
of the $2{,}995$ validation frames; each frame carries one tag per category, so
rows of different categories overlap. Each subset is scored with the official
metrics restricted to its frames. Seed-0 and seed-1 \method{} predictions are
those of Table~\ref{tab:review}.}
\label{tab:conditions}
\begin{tabular}{ll r rrr r}
\toprule
condition & & frames & seed 0 & seed 1 & seed 2 & mean \\
\midrule
area & city & $320$ & $+1.38$ & $+3.47$ & $+0.92$ & $+1.92$ \\
 & highway & $1{,}960$ & $+0.11$ & $+2.46$ & $+1.22$ & $+1.26$ \\
 & rural & $315$ & $+0.46$ & $+0.89$ & $-1.68$ & $-0.11$ \\
 & terminal & $320$ & $-0.13$ & $+0.83$ & $-0.21$ & $+0.16$ \\
\addlinespace
weather & clear & $1{,}560$ & $+1.82$ & $+2.50$ & $-1.72$ & $+0.87$ \\
 & overcast & $915$ & $-1.03$ & $+1.85$ & $+1.22$ & $+0.68$ \\
 & rain & $400$ & $-0.37$ & $+0.50$ & $+0.35$ & $+0.16$ \\
\addlinespace
lighting & dark & $360$ & $+2.01$ & $-0.48$ & $-0.14$ & $+0.46$ \\
 & illuminated & $1{,}955$ & $+0.89$ & $+2.54$ & $+0.09$ & $+1.17$ \\
 & twilight & $440$ & $+1.72$ & $+0.44$ & $+0.59$ & $+0.92$ \\
\addlinespace
daytime & evening & $480$ & $+3.92$ & $-0.99$ & $+0.67$ & $+1.20$ \\
 & morning & $1{,}000$ & $+0.22$ & $+3.70$ & $-1.26$ & $+0.89$ \\
 & noon & $1{,}515$ & $+1.32$ & $+2.45$ & $-0.45$ & $+1.11$ \\
\addlinespace
season & autumn & $1{,}360$ & $+2.05$ & $+2.50$ & $+0.23$ & $+1.59$ \\
 & summer & $880$ & $+0.38$ & $+3.22$ & $-0.95$ & $+0.88$ \\
 & winter & $755$ & $+1.28$ & $+1.15$ & $+0.55$ & $+0.99$ \\
\addlinespace
structure & overpass & $360$ & $+1.65$ & $+3.14$ & $-0.06$ & $+1.58$ \\
 & regular & $1{,}995$ & $+0.61$ & $+4.37$ & $+0.30$ & $+1.76$ \\
 & underpass & $480$ & $-3.85$ & $+1.72$ & $+2.34$ & $+0.07$ \\
\addlinespace
construction & unchanged & $2{,}875$ & $+1.63$ & $+2.18$ & $-0.16$ & $+1.22$ \\
\bottomrule
\end{tabular}
\end{table}

\begin{figure}[htbp]\centering
\includegraphics[width=\linewidth]{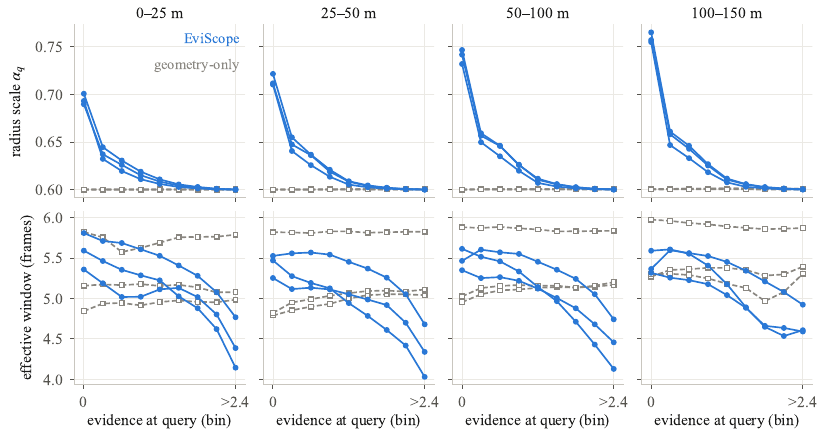}
\caption{Allocation with and without the evidence input, inside each
range band. Probe of Table~\ref{tab:probe} on \method{} (blue, filled) and the
geometry-only control (grey, hollow), seeds $0$--$2$ of each. Queries are
binned by the evidence read at the query: the first bin holds zero readings,
the other eight are quantile bins of the positive readings pooled over all
models. Top: mean radius scale $\alpha_q$; bottom: mean effective window
$\exp(\mathrm{H})$ of the frame weights, in frames. Bins with fewer than $50$
queries are omitted. The control's radius is $0.600$--$0.607$ for every
query.}
\label{fig:probebins}
\end{figure}

\textbf{What the control learns.} We ran the probe of
Table~\ref{tab:probe} (last decoder iteration, first $200$ validation samples,
$36{,}000$ queries per model) on the three control models and on \method{}
seeds $0$--$4$, and computed rank correlations within each range band.
\method{}'s radius falls as the evidence at the query rises, in every band and
seed (Spearman $-0.89$ to $-0.95$), and its effective window shortens ($-0.06$
to $-0.73$, negative in all $20$ seed-by-band cells). The control's radius
stays at the branch's lower bound, $0.600$--$0.607$ for every query (standard
deviation below $0.001$ at each seed), and its window has no consistent
relation to the evidence ($-0.26$ to $+0.37$). Without the evidence input,
every query samples as narrowly as \method{} samples its best-observed queries
(Figure~\ref{fig:probebins}).

\textbf{Eight-epoch result.} The seed-1 pair, trained and evaluated on one
node, gives \method{} $26.49$ / $40.08$ and the baseline $26.62$ / $38.89$:
$+1.19$ NDS and $-0.13$ mAP. With the seed-0 pair of Table~\ref{tab:budget}
($+0.67$ NDS, $+0.26$ mAP), \method{} is above its baseline on NDS in both
completed eight-epoch pairs, by $+0.93\pm0.26$ on average (mean $\pm$ SE), with
mAP level ($+0.07\pm0.20$). At seed 1 the NDS difference comes from orientation,
velocity and scale error (mAOE $0.548\to0.457$, mAVE $0.743\to0.719$, mASE
$0.420\to0.407$); translation error is similar ($0.603$ and $0.602$).
These two pairs do not establish an improvement with longer training.

\textbf{Density-head result.} Over seeds $0$--$2$, \method{} exceeds the
same module trained without the density branch by $+0.67\pm0.68$ NDS and
$+0.11\pm0.25$ mAP (mean $\pm$ SE). The NDS differences are $+1.26$, $+1.43$
and $-0.69$, and the mAP differences $+0.01$, $+0.59$ and $-0.26$. Against the
baseline, the no-density variant averages $+0.48\pm0.31$ NDS and $-0.03\pm0.35$
mAP; it is also above the geometry-only control in all three pairs (mAP $+1.11$,
$+0.45$, $+0.72$; NDS $+0.36$, $+0.83$, $+0.55$; evaluated on different nodes at
seeds $0$ and $1$), noting that those two models differ in both the evidence
input and the auxiliary loss.

\textbf{Shuffled-evidence result.} Permuting the evidence channels across
the queries of each sample gives $25.19$ / $37.30$ at seed $0$ and $26.31$ /
$38.25$ at seed $1$ (mAP / NDS), so \method{} exceeds the shuffled control in
both completed pairs: $+1.28$ mAP / $+2.89$ NDS at seed $0$ and $+0.20$ mAP /
$+1.99$ NDS at seed $1$, averaging $+0.74\pm0.54$ mAP and $+2.44\pm0.45$ NDS
($+0.80\pm0.52$ mAP and $+2.47\pm0.43$ NDS when evaluated on the control's
node). Translation, orientation and velocity errors are higher in the shuffled
control at both seeds (mATE $0.664$ vs.\ $0.636$ and $0.652$ vs.\ $0.624$;
mAOE $0.578$ vs.\ $0.473$ and $0.567$ vs.\ $0.499$; mAVE $0.740$ vs.\ $0.665$
and $0.720$ vs.\ $0.641$), and both shuffled runs also fall below their
matched baselines ($25.84$ / $38.48$ and $26.48$ / $38.85$).
The ordering against geometry-only is mixed: shuffled minus geometry-only is
$-0.16$ / $-1.27$ at seed $0$ and $+0.84$ / $+0.27$ at seed $1$ (mAP / NDS).
These two pairs support the value of spatial alignment under this control,
without establishing an ordering between all control variants.

\clearpage
\section{Settings of every comparison}
\label{app:settings}
Table~\ref{tab:settings} lists, for every comparison in the paper, the
schedule, the number of runs and what differs between its rows. All training
runs use one node of eight A100 GPUs with a per-GPU batch of one, ResNet-50 at
$256\times704$ unless stated, and the official TruckScenes evaluator.

\begin{table}[htbp]\centering\scriptsize
\setlength{\tabcolsep}{3pt}
\caption{Summary of experimental settings for the comparisons in this paper (Table~\ref{tab:review},
Table~\ref{tab:conditions}, Figure~\ref{fig:probebins}). \emph{Runs} counts
completed trained models per arm used in the reported comparisons.
Inference-time substitutions and probes re-evaluate trained weights without retraining.}
\label{tab:settings}
\begin{tabular}{p{0.25\linewidth}p{0.15\linewidth}p{0.15\linewidth}p{0.35\linewidth}}
\toprule
comparison & schedule & runs & what differs \\
\midrule
Seven matched pairs (Tab.~\ref{tab:seeds}) & 4 ep & 7 + 7 & module on / off, same seed \\
Eight epochs (Tab.~\ref{tab:budget}, \ref{tab:review}) & 8 ep & 2 + 2 & module on / off, same seed \\
Geometry-only (Tab.~\ref{tab:review}) & 4 ep & 3 & field channels zeroed in $d_q$ \\
Shuffled evidence (Tab.~\ref{tab:review}) & 4 ep & 2 & field channels permuted across queries of each sample \\
Density head (Tab.~\ref{tab:ablation4}, \ref{tab:review}) & 4 ep & 3 & density branch removed \\
Higher resolution (Tab.~\ref{tab:main}) & 8 ep & 1 & $256\times864$; no paired baseline \\
Component ablation (Tab.~\ref{tab:ablation4}) & 4 ep & 1 each & cumulative branches; fixed radius, no module \\
Single-epoch suite (Tab.~\ref{tab:ablation}) & 1 ep, multi-frame from step 1 & 1 each & branches, density head, evidence input \\
Recipe (Tab.~\ref{tab:recipe}) & 4 ep (12 ep without resampling) & 1 each & resampling, frame count \\
V2-99 pair (Tab.~\ref{tab:v299}) & 4 ep & 1 + 1 & backbone \\
Substitutions (Tab.~\ref{tab:channel}) & trained weights & 4 \method{} + 2 controls & one modulation replaced at inference \\
Held-out recall (Tab.~\ref{tab:heldout}) & 4 ep & seeds 5, 6 & module on / off; subset fixed from seeds 0--4 \\
Probe (Tab.~\ref{tab:probe}) & trained weights & 4 & none (hooks only) \\
Scene conditions (Tab.~\ref{tab:conditions}) & trained weights & 3 + 3 & official scene tags, $\ge 300$ frames \\
Probe on the control (Fig.~\ref{fig:probebins}) & trained weights & 5 + 3 & none (hooks only) \\
Degradation (Tab.~\ref{tab:degrade}) & trained weights & 1 + 1 & radar inputs dropped, same mask for both \\
\bottomrule
\end{tabular}
\end{table}

\end{document}

%% file: authors.tex
\begin{minipage}[t]{0.96\textwidth}\normalfont\centering\fontsize{10.5}{13}\selectfont
\mbox{Mingyang Lyu$^{1,2,3}$}\quad \mbox{Yinqian Sun$^{1}$}\quad \mbox{Gongce Cao$^{3}$}\\[2pt]\mbox{Chujie Tang$^{1}$}\quad \mbox{Yuhao Zhang$^{4}$}\quad \mbox{Lu Jia$^{5}$}\\[2pt]\mbox{Qian Zhang$^{1}$}\quad \mbox{Bo Wang$^{1,\dagger}$}\quad \mbox{Pin Lyu$^{1,\dagger}$}
\\[5pt]{\fontsize{8.5}{10.5}\selectfont 
$^{1}$Institute of Automation, Chinese Academy of Sciences, Beijing, China\\$^{2}$Zhongguancun Academy, Beijing, China\\$^{3}$University of Chinese Academy of Sciences, Beijing, China\\$^{4}$University of Illinois Urbana-Champaign, USA\\$^{5}$University of Science and Technology Beijing, Beijing, China
\\[4pt]$^\dagger$Corresponding authors: Pin Lyu, Bo Wang.}
\end{minipage}